\documentclass[10pt,twocolumn,letterpaper]{article}

\usepackage{iccv}      
\iccvfinalcopy
\usepackage{times}

\usepackage{graphicx}
\usepackage{amsmath}
\usepackage{amssymb}
\usepackage{booktabs}
\usepackage{multicol}
\usepackage{lipsum}
\usepackage{subcaption}
\usepackage{colortbl}
\usepackage{xcolor}
\usepackage{caption}
\usepackage{booktabs}
\usepackage{overpic}
\usepackage{array}
\usepackage{pifont}
\usepackage[toc,page]{appendix}

\usepackage[pagebackref=true,breaklinks=true,letterpaper=true,colorlinks,bookmarks=false]{hyperref}

\def\iccvPaperID{5777} 

\ificcvfinal\fi
\begin{document}


\title{SQLdepth: Generalizable Self-Supervised Fine-Structured Monocular Depth Estimation}

\author{Youhong Wang$^{1, 2}$, Yunji Liang$^{1}$, Hao Xu$^{2}$, 
Shaohui Jiao$^{2}$, Hongkai Yu$^3$ \\
\@\textsuperscript{1}Northwestern Polytechnical University, \@\textsuperscript{2}Bytedance Inc, \@\textsuperscript{3}Cleveland State University
}
\maketitle

\ificcvfinal\thispagestyle{empty}\fi
\begin{abstract}
  Recently, self-supervised monocular depth estimation has gained popularity
  with numerous applications in autonomous driving and robotics. However,
  existing solutions primarily seek to estimate depth from immediate visual
  features, and struggle to recover fine-grained scene details with limited
  generalization. In this paper, we introduce SQLdepth, a novel approach that
  can effectively learn fine-grained scene structures from motion. In SQLdepth,
  we propose a novel Self Query Layer (SQL) to build a self-cost volume and
  infer depth from it, rather than inferring depth from feature maps. 
  The self-cost volume implicitly captures the intrinsic geometry of the scene
  within a single frame. Each individual slice of the volume signifies the
  relative distances between points and objects within a latent space.
  Ultimately, this volume is compressed to the depth map via a novel decoding approach.
  Experimental results on KITTI and Cityscapes show that our method attains
  remarkable state-of-the-art performance (AbsRel = $0.082$ on KITTI, $0.052$
  on KITTI with improved ground-truth and $0.106$ on Cityscapes), achieves
  $9.9\%$, $5.5\%$ and $4.5\%$ error reduction from the previous best. 
  In addition, our approach showcases reduced training
  complexity, computational efficiency, improved generalization, and the ability to recover fine-grained scene details.
  Moreover, the self-supervised pre-trained and metric fine-tuned SQLdepth 
  can surpass existing supervised methods by significant margins (AbsRel = $0.043$, $14\%$ error reduction).
  Code is available at \href{https://github.com/hisfog/SQLdepth-Impl}{https://github.com/hisfog/SQLdepth-Impl}.
 
\end{abstract}

\section{Introduction}
\label{sec:intro}

Monocular depth estimation is a fundamental research topic in
computer vision and is widely used in numerous applications, such as autonomous
driving \cite{Geiger2013IJRR}, 3D reconstruction
\cite{RichardNewcombe2011KinectFusionRD}, augmented reality
\cite{XuanLuo2020ConsistentVD} and robotics \cite{dudek2010computational,
achtelik2009stereo}. Given a single RGB image, monocular depth estimation aims
to predict the corresponding depth value of each pixel. For the supervised
methods, sparse depth ground truth collected by sensors such as LiDAR is
required. However, it is time-consuming and expensive to collect large-scale
depth information from physical world. In addition, supervised depth estimators
can not be well-optimized under the sparse supervision and has limited
generalization to unseen scenarios.

\begin{figure}[t]
  \setlength{\columnsep}{0.5pt} 
\begin{multicols}{2}
  \centering

  \begin{subfigure}[b]{0.25\textwidth}
    \begin{minipage}[t]{\textwidth}
      \centering
      \begin{overpic}[width=0.95\linewidth,keepaspectratio]{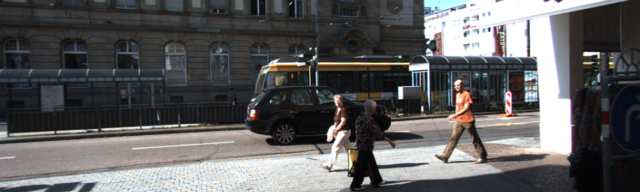}
        \put(82,2){RGB}
      \end{overpic}
      \begin{overpic}[width=0.95\linewidth,keepaspectratio]{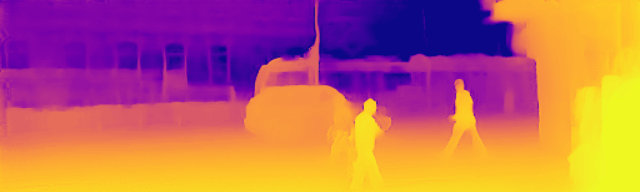}
        \put(82,2){Ours}
      \end{overpic}
      \begin{overpic}[width=0.95\linewidth,keepaspectratio]{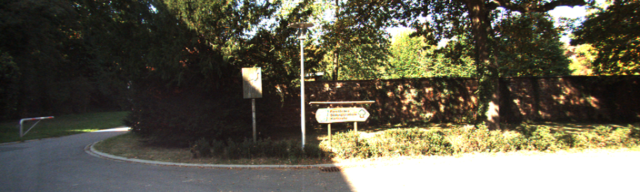}
        \put(82,2){RGB}
      \end{overpic}
      \begin{overpic}[width=0.95\linewidth,keepaspectratio]{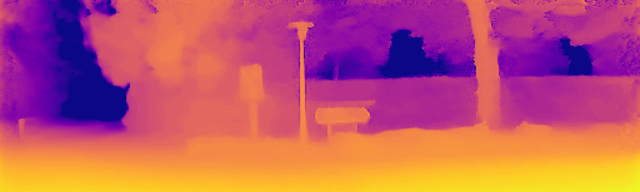}
        \put(82,2){Ours}
      \end{overpic}

      \begin{overpic}[width=0.95\linewidth,keepaspectratio]{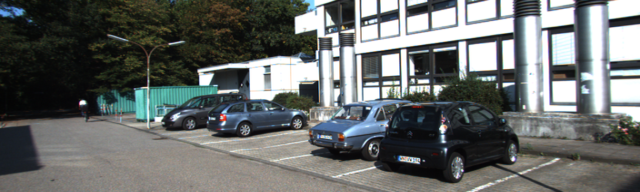}
        \put(82,2){RGB}
      \end{overpic}
      \begin{overpic}[width=0.95\linewidth,keepaspectratio]{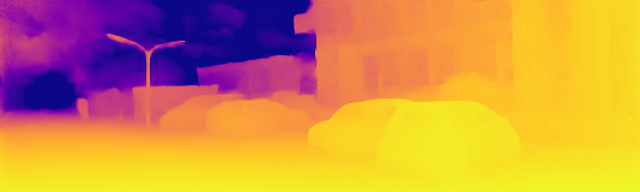}
        \put(82,2){Ours}
      \end{overpic}
      \begin{overpic}[width=0.95\linewidth,keepaspectratio]{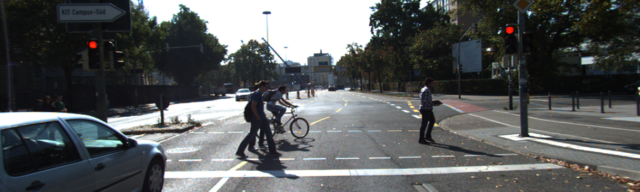}
        \put(82,2){RGB}
      \end{overpic}
      \begin{overpic}[width=0.95\linewidth,keepaspectratio]{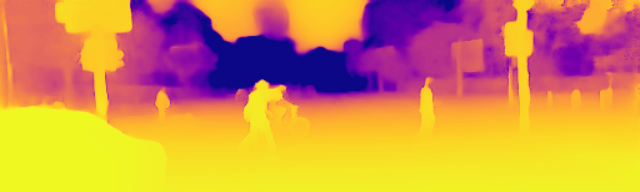}
        \put(82,2){Ours}
      \end{overpic}

    \end{minipage}%
  \end{subfigure}

  \begin{subfigure}[b]{0.25\textwidth}
    \tiny
    \begin{minipage}[t]{\textwidth}
      \centering
      \begin{overpic}[width=0.95\linewidth,keepaspectratio]{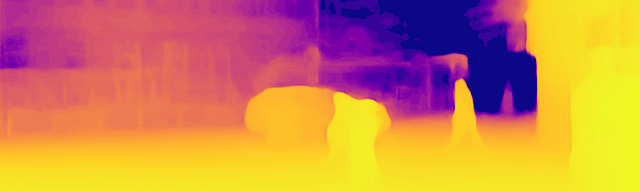}
        \put(70,1){Monodepth2 \cite{godard2019digging}}
      \end{overpic}
      \begin{overpic}[width=0.95\linewidth,keepaspectratio]{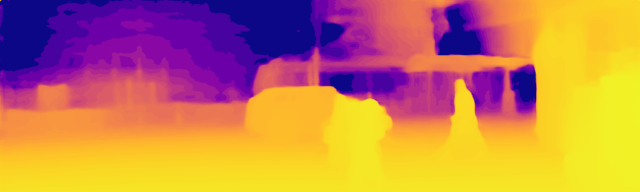}
        \put(73,2){EPCDepth \cite{peng2021excavating}}
      \end{overpic}

      \begin{overpic}[width=0.95\linewidth,keepaspectratio]{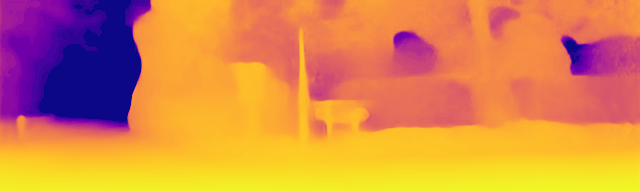}
        \put(70,1){Monodepth2 \cite{godard2019digging}}
      \end{overpic}
      \begin{overpic}[width=0.95\linewidth,keepaspectratio]{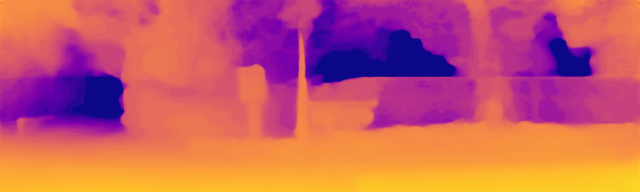}
        \put(73,2){EPCDepth \cite{peng2021excavating}}
      \end{overpic}

      \begin{overpic}[width=0.95\linewidth,keepaspectratio]{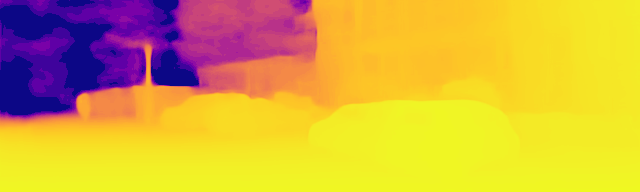}
        \put(70,1){Monodepth2 \cite{godard2019digging}}
      \end{overpic}
      \begin{overpic}[width=0.95\linewidth,keepaspectratio]{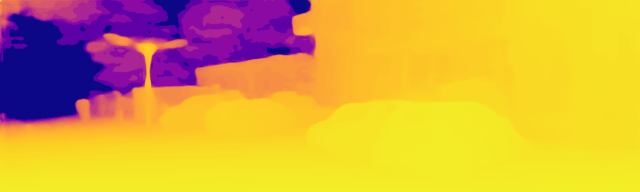}
        \put(73,2){EPCDepth \cite{peng2021excavating}}
      \end{overpic}
      \begin{overpic}[width=0.95\linewidth,keepaspectratio]{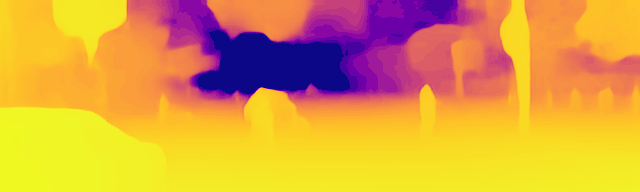}
        \put(70,1){Monodepth2 \cite{godard2019digging}}
      \end{overpic}
      \begin{overpic}[width=0.95\linewidth,keepaspectratio]{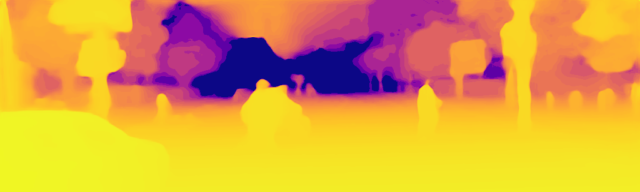}
        \put(73,2){EPCDepth \cite{peng2021excavating}}
      \end{overpic}

    \end{minipage}%
  \end{subfigure}

\end{multicols}
  \caption{\textbf{Typical examples of our predictions on images from KITTI
  \cite{Geiger2013IJRR} dataset.} 
  Compared with the classical method
  Monodepth2 \cite{godard2019digging} and the previous best EPCDepth
  \cite{peng2021excavating}, our method is able to recover much more scene details,
  such as traffic signs, street lamps and pedestrians.
  }
  \label{img:first}
\end{figure}

Lately, self-supervised solutions have gained popularity. 
Existing efforts have concentrated on training with self-distillation
\cite{peng2021excavating}, leveraging depth hints \cite{watson2019self},
and infering with multi-frames \cite{watson2021temporal, feng2022disentangling}.
However, they often fail to recover fine-grained scene details,
as shown in Figure. \ref{img:first}. 
How to learn the fine-grained scene structures effectively and
efficiently in self-supervised setting is still challenging.

In this paper, we propose to estimate
depth from our carefully designed self-cost volume that stores relative
distance representations rather than estimate depth from immediate visual
features. The motivation of our work is based on the following 
observation: the depth of a pixel depth is strongly correlated with the depth 
of its adjacent pixels and related objects within the image \cite{yuan2022new, BTS}. 
This suggests that a pixel's depth can be inferred from related contexts, which
provide \textit{relative} distance information. 

Consequently, to build the self-cost volume, we first model points and objects in a latent space.
Specifically, we employ a Convolutional Neural Network (CNN) to extract point
features and a compact Vision Transformer (ViT) to extract object queries. 
Secondly, within a novel Self Query Layer (SQL), 
we use feature dot-product to compare each
pixel with each object to build the self-cost volume.
Hence, each slice of the volume indicates the related distance map provided by a specific object query. 
Finally, we propose an novel and effective decoding approach 
specifically tailored for compressing the self-cost volume to the final depth map.

Our main contributions are as follows:
\begin{itemize}
    \item 
      Introducing SQLdepth, a novel self-supervised method empowered by the Self
      Query Layer (SQL) to construct a self-cost volume that effectively
      captures fine-grained scene geometry of a single image.
  
    \item 
      Demonstrating through comprehensive experiments on KITTI and Cityscapes
      datasets that SQLdepth is simple yet effective,
      and surpasses existing self-supervised alternatives in
      accuracy and efficiency.
    
    \item 
    Demonstrating SQLdepth's improved generalization. 
    by applying a
    KITTI pre-trained model to other datasets, such as zero-shot transfer to Make3D.
\end{itemize}

\section{Related works}

\subsection{Supervised Depth Estimation}
Eigen \textit{et al.} \cite{eigen2014depth} was the first to propose a
learning based approach which
utilizing a multiscale convolutional neural network as well as a scale-invariant loss function
for predicting depth from a single image.
Afterwards, numerous approaches have been proposed. 
Generally, these methods can be categorized into two groups:
methods formulating the depth estimation task as 
a pixel-wise regression
problem \cite{eigen2014depth,huynh2020guiding, ranftl2021vision,
zhao2021transformer}, or formulating it as a pixel-wise classification problem \cite{fu2018deep,
diaz2019soft}. 
The regression based methods can predict continuous depths, but are hard to optimize.
The classification based methods can only predict discrete depths, but are easier to optimize.

To combine the strengths of both regression and classification tasks,
studies in \cite{ShariqFarooqBhat2020AdaBinsDE, johnston2020self} reformulate
depth estimation as a per-pixel classification-regression task. 
They propose to first regress a set of depth bins
and then perform pixel-wise classification to assign each pixel to the corresponding bin.
The final depth is a linear combination of bin centers weighted by the probabilities.
This approach has attains remarkable improvement in precision.

\subsection{Self-supervised Depth Estimation}

In the absence of ground truth, self-supervised models are usually trained by
either making use of the 
temporal scene consistency in monocular videos \cite{2017Unsupervised, godard2019digging},
or left-right scene consistency in stereo image
pairs \cite{garg2016unsupervised, godard2017unsupervised, pillai2019superdepth}.

\noindent\textbf{Monocular Training.} 
In monocular training, supervision comes from the consistency between 
the synthesis scene view from referenced frame and the scene view of source frame.
SfMLearner \cite{2017Unsupervised} jointly trained a DepthNet and a separate PoseNet
under the supervision of a photometric loss. 
Following this classical joint-training pipeline, many advances were
proposed to improve the learning process, e.g. more robust reconstruction
loss of image level \cite{gordon2019depth, shu2020feature}, feature level
reconstruction loss \cite{shu2020feature, zhan2018unsupervised},
using auxiliary information during training \cite{watson2019self, MariaKlodt2018SupervisingTN}, 
dealing with moving objects that break the assumption of static scene during
training \cite{AnuragRanjan2018CompetitiveCJ, godard2019digging,
gordon2019depth, SudheendraVijayanarasimhan2017SfMNetLO,
ZhichaoYin2018GeoNetUL, YuhuaChen2019SelfSupervisedLW,
JiaWangBian2019UnsupervisedSD, VincentCasser2019UnsupervisedMD,
MarvinKlingner2020SelfSupervisedMD, HanhanLi2020UnsupervisedMD}, and introducing
extra constraints \cite{ZhenhengYang2018UnsupervisedLO, ZhichaoYin2018GeoNetUL,
ranjan2019competitive, PoYiChen2019TowardsSU,
VitorGuizilini2020SemanticallyGuidedRL, ShengjieZhu2020TheEO, PLADENet}.

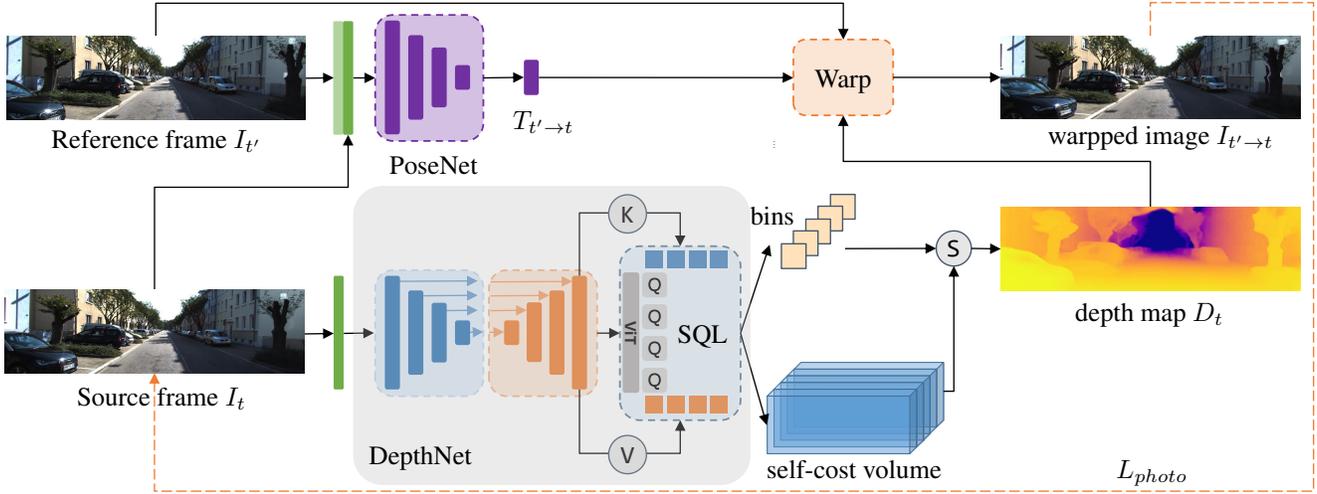
\begin{figure*}[t]
 \centering
 \begin{minipage}{\textwidth}
 \centering
      \begin{overpic}[scale=2.47,keepaspectratio]{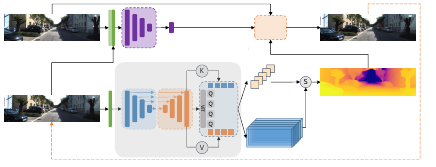}
        \put(82,2.5){$L_{photo}$}
        \put(6.2,7.8){Source frame $I_t$}
        \put(27.5,3.5){DepthNet}
        \put(50,12.4){SQL}
        \put(29,24.7){PoseNet}
        \put(38,28){$T_{t^\prime \rightarrow t}$}
        \put(55.3,21){bins}
        \put(60,31){Warp}
        \put(4.35,26.6){Reference frame $I_{t^\prime}$}
        \put(77,26.9){warpped image $I_{t^\prime \rightarrow t}$}
        \put(79,14.0){depth map $D_t$}
        \put(56.5,2.5){self-cost volume}
      \end{overpic}
   \captionof{figure}{
   \textbf{Framework Overview:} \textbf{(1) DepthNet:} We
   first use a standard convolutional encoder-decoder to extract immediate
   visual representations of frame $I_t$, and then feed these visual features
   into Self Query Layer (See Figure \ref{img:SQL layer} for more details) 
   to obtain the depth map $D_t$ of current frame.
   \textbf{(2) PoseNet:} Relative pose between the current frame
   $I_t$ and the reference frame $I_{t^{\prime}}$ is predicted with a 
   standard pose network. The camera pose is only needed during training 
   to perform differentiable warpping. 
   \textbf{(3) Differentiable warpping:} Following many previous works,
   we use pixels in
   frame $I_{t^{\prime}}$ to reconstruct frame $I_t$, with the help of depth
   map $D_t$ and relative pose $T_{t^\prime \rightarrow t}$, by the
   differentiable warpping process \cite{jaderberg2015spatial}. 
   The loss function is built upon the differences of the warpped image $I_{t^\prime \rightarrow t}$ and
   source image $I_t$.}

 \label{fig:overview}
 \end{minipage}
\end{figure*}
\noindent\textbf{Stereo Training.} 
Stereo training uses synchronized stereo pairs of images and predicts the disparity map \cite{DanielScharstein2001ATA},
which means the inverse of depth map.
In the setting of stereo training, the relative camera pose is known,
and the model only needs to predict the disparity map. 
Garg \textit{et al.}  \cite{garg2016unsupervised} trained a self-supervised
monocular depth estimator with photometric consistency loss between stereo pairs. 
Following this, more constraints were proposed to optimize the network,
including left-right consistency \cite{godard2017unsupervised}, temporal
consistency in videos \cite{zhan2018unsupervised}.
Garg \textit{et al.}  \cite{garg2020wasserstein} 
extended this approach by predicting continuous disparity values.
Stereo-based approaches have been further extended
with semi-supervised data  \cite{kuznietsov2017semi, YueLuo2018SingleVS},
using auxiliary information \cite{watson2019self}, using exponential probability volumes \cite{MED_NIPS},
and using self-distillation mechanisms \cite{peng2021excavating, guo2018learning, AndreaPilzer2019RefineAD}.
Generally, stereo views can serve as perfect reference frames providing valuable supervision,
or can also be used to obtain the absolute depth scale.

However, all existing self-supervised methods fail to produce high-quality depth maps.
In fact, existing methods are either estimating depth from immediate visual features,
or from Transformer \cite{dosovitskiy2020image} enhanced high-level visual representations.
They overlook the importance of pixel-level geometric cues that may be constructive to model
performance as well as generalization capacity.




\section{Problem Setup}

The goal of self-supervised monocular depth estimation is to predict the depth
map from a single RGB image without ground truth, which can also be viewed as
learning structure from motion (SfM). 
As illustrated in Figure \ref{fig:overview},
given a single source image $I_t$ as input, first, the DepthNet predicts its corresponding depth map
$D_t$. And the PoseNet takes both source
image and reference image ($I_t$, $I_{t^{\prime}}$) as input and
predicts the relative pose $T_{t \rightarrow t^\prime}$ between the source image
$I_t$ and reference image $I_{t^{\prime}}$. 
Finally, we use the predicted $D_t$ and $T_{t \rightarrow t^\prime}$ to perform view
synthesis by Eq. \ref{synthesis}.
\begin{equation}
  \label{synthesis}
I_{t^\prime \rightarrow t}=I_{t^\prime}\left\langle\operatorname{proj}\left(D_t, T_{t \rightarrow t^\prime}, K\right)\right\rangle 
\end{equation}
where $\langle\rangle$ is the sampling operator and $\operatorname{proj}$ returns the 2D
coordinates of the depths in $D_t$ when reprojected into the camera view of
$I_{t^\prime}$. 
At training time, both the DepthNet and PoseNet are optimized jointly
by minimizing the photometric reprojection error.
Following \cite{garg2016unsupervised, 2017Unsupervised, zhou2017unsupervised}, for each pixel we
optimize the loss for the best matching source image, by selecting the
per-pixel minimum over the reconstruction loss $pe$ defined in Eq. \ref{769},
where $t^\prime \in (t-1, t+1)$.
More details about $pe$ are provided in Section \ref{sec:pe}. 
\begin{equation}
  \label{769}
L_p=\min _{t^\prime} p e\left(I_t, I_{t^\prime \rightarrow t}\right) 
\end{equation}

\section{Method}
In this section, we elaborate the design details of the two 
 core components in the DepthNet: the CNN-based encoder-decoder for extracting immediate visual representations, and the Transformer-based Self Query Layer (SQL) for depth decoding.
\label{sec:formatting}
\begin{figure}[ht]
 \centering
  \centering
   \includegraphics[width=1.05\linewidth]{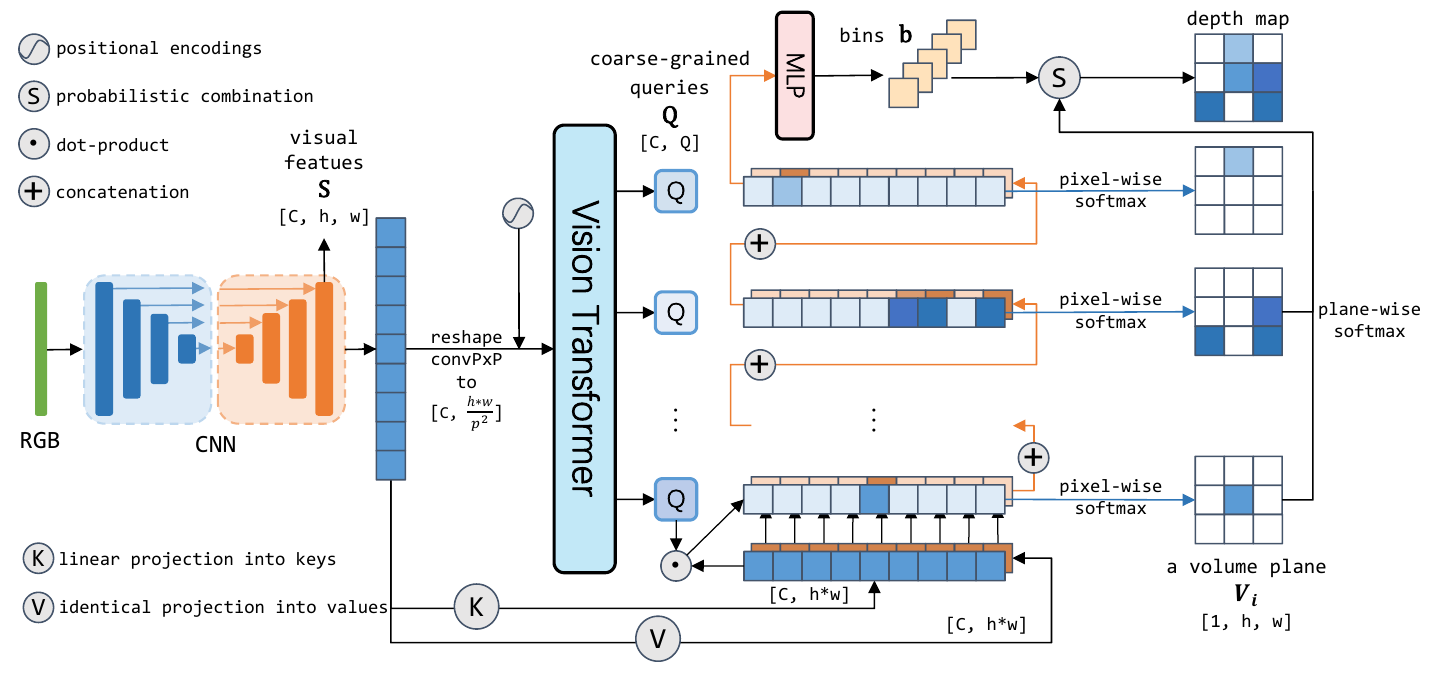}

   \caption{
     Framework details of DepthNet with the proposed Self Query Layer.
   }
   \label{img:SQL layer}
\end{figure}

\subsection{Extract Immediate Visual Representations}
\textcolor{black}{Given an input RGB image of shape $\mathbb{R}^{3 \times H \times W}$, the CNN-based encoder-decoder first extracts
image features and decodes with upsampling them into high resolution immediate
features $\mathbf{S}$ of shape $\mathbb{R}^{C \times h \times w}$.
Considering of the hardware limitation, we set $h=\frac{H}{2}, w=\frac{W}{2}$.
Benefiting from the encoder-decoder framework with
skip connections, we can extract useful local fine-grained visual clues.}

\subsection{Self Query Layer}\label{sec:SQL}
 
\noindent\textbf{Building a self-cost volume $\mathbf{V}$.}
As one of the geometric clues, relative distance information is beneficial for the geometric depth estimation task.
But how to obtain the relative distance remains as a problem.
Inspired by previous works \cite{watson2021temporal, teed2020raft},
where cost volume is used to capture cross image geometric clues for other
  geometric tasks (MVS \cite{watson2021temporal} and optical flow
  \cite{teed2020raft}), we can define the dot-product value of two pixel
  embeddings in a single image to capture the intra-geometric clues (relative
  distance representations in a single image) for depth estimation, and store
  it in a self-cost volume. However, the time complexity of this procedure is
  $O(h^2 \times w^2)$,
which makes it infeasible to build the self-cost volume directly upon the
  feature map $\mathbf{S}$. Therefore, we propose to calculate the relative
  distance approximately. Concretely, we can define the relative distance as
  the dot product between a pixel and patch, and store it in a self-cost volume
  for later usage and the overall computational complexity is reduced to $O(h
  \times w)$.

\noindent\textbf{Coarse-grained queries $\mathbf{Q}$.}
\textcolor{black}{The above self-cost volume represents the pixel-patch relative
  distance but in fact the distance is lack of physical significance.
  Actually, we can obtain the relative distance
  between pixels and objects, just by one more step. 
  To achieve this, we first introduce the 
  coarse-grained queries to represent objects in the image.
  Concretely, we can utilize the dynamic receptive
  field of transformers to enhance the patch embeddings, enabling them to
  represent objects in the image implicitly. 
  And after enhancing, we use these coarse-grained queries
  to do per-pixel relative distance querying (dot product) to get the
  pixel-level geometric clues.} 

Specifically, we first apply a convolution of kernel size $p \times p$ and stride $=p$ (e.g. $p = 16$)
to $\mathbf{S}$, getting a feature map $\mathbf{F}$ of shape
$C \times \frac{h}{p} \times \frac{w}{p}$.
Second, we reshape $\mathbf{F}$ to $\mathbb{R}^{C \times N}$, 
and add positional embeddings to $\mathbf{F}$,
where $N=\frac{h*w}{p*p}$ is the number of patches.
Then we feed these patch embeddings into a
mini-transformer of $4$ layers to generate a set of coarse-grained queries $\mathbf{Q}$ of shape $\mathbb{R}^{C \times Q}$,
where $Q$ is a hyperparameter and $Q \leq N$.
Finally,
we apply these coarse-grained queries to per-pixel immediate visual representations in $\mathbf{S}$ to get 
the self-cost volume $\mathbf{V}$ of size $h \times w \times Q$, 
where $V_{i, j, k}$ is calculated as Eq. \ref{volume}.

\begin{equation}
  \label{volume}
V_{i, j, k}=Q_i^T \cdot S_{j, k} 
\end{equation}

\noindent\textbf{Depth bins estimation with the self-cost volume.}
Prior works \cite{ShariqFarooqBhat2020AdaBinsDE, binsFormer} 
show that depth bins are useful for estimating continuous depth.
Bhat \textit{et. al} \cite{ShariqFarooqBhat2020AdaBinsDE} 
employs a brute force regression method
to calculate depth bins from a token extracted by a Vision Transformer \cite{dosovitskiy2020image},
and utilizes a chamfer loss as supervision.
However, this brute force approach fails in self-supervised setting, see ablation in Table \ref{ablation:approximation}.
Therefore, we rethink the essence of depth bins.
Depth bins essentially represent the distribution of depth,
that is, the countings of different depth values.
Therefore, we propose to estimate depth bins $\mathbf{b}$ 
by counting the latent depths in the self-cost volume.
Since we perform in a latent depth space,
we can view the counting process as a information aggregation process,
and use two basic operations for information aggregation: softmax and weighted sum to achieve the counting operation.
Specifically, 
for every plane in the self-cost volume $\mathbf{V}$,
we first apply a pixel-wise softmax 
to convert the volume-plane into a pixel-wise probabilistic map.
Then we perform a weighted sum of per-pixel visual representations in $\mathbf{S}$ using this map. 
After this procedure, we get $Q$ vectors of dimension $C$, representing $Q$ depth countings in $Q$ planes.
Finally we concat them and feed it into a MLP to regress the depth bins $\mathbf{b}$ as shown in Eq. \ref{eq:bins}:
\begin{equation}
\label{eq:bins}
\mathbf{b}=MLP\left(\bigoplus_{i=1}^Q \sum_{(j,k)=(1,1)}^{(h,w)}softmax(V_i)_{j,k} \cdot S_{j,k}\right)
\end{equation}

\noindent\textbf{Probabilistic combination.}
From above we have employed a set of coarse-grained queries to do pixel-wise querying, each
of which represents different image contexts and produces its own depth estimation.
In this paragraph, we will combine all these depth estimations to 
get the final depth map. Firstly, in order to match the dimension of depth bins $\mathbf{b}$
of shape $D$, we apply a $1 \times 1$ convolution to the self-cost volume
$\mathbf{V}$ to obtain a $D$-planes volume.
Secondly, we apply a plane-wise softmax operation to convert the volume into plane-wise probabilistic maps
as shown in Eq. \ref{eq:5}.
\begin{equation}
\label{eq:5}
p_{i,j,k}=softmax(V)_{i,j,k}, 1 \leq i \leq Q
\end{equation}
\textcolor{black}{Finally, for every pixel, the depth is calculated by
aggregating the centers of the bins in the standard probabilistic linear
combination manner \cite{ShariqFarooqBhat2020AdaBinsDE} as shown in Eq.
\ref{eq:6}:}
\begin{equation}
\label{eq:6}
\tilde{d}=\sum_{i=1}^N c\left(b_i\right) p_{i,j,k}, 1 \leq j \leq h, 1 \leq k \leq w
\end{equation}
\textcolor{black}{where $c\left(b_i\right)$ is the center depth of the $i^{th}$
bin and it is determined by Eq~\ref{interval}. }
\begin{equation}
  \label{interval}
c\left(b_i\right)=d_{\min }+\left(d_{\max }-d_{\min }\right)\left(b_i / 2+\sum_{j=1}^{i-1} b_j\right) 
\end{equation}
where $\mathbf{b}$ denotes the depth distribution bins we calculated.

\begin{table*}[t]
\centering
  \resizebox{\textwidth}{!}{ 
  \begin{tabular}{|c|c|c|c||c|c|c|c|c|c|c|}
\hline
  Method &Train &Test &HxW &\cellcolor{red!25}$AbsRel\downarrow$ &\cellcolor{red!25}$SqRel\downarrow$ &\cellcolor{red!25}$RMSE\downarrow$ &\cellcolor{red!25}$RMESlog\downarrow$ &\cellcolor{blue!25}$\delta<1.25\uparrow$ &\cellcolor{blue!25}$\delta<1.25^2\uparrow$ &\cellcolor{blue!25}$\delta<1.25^3\uparrow$   \\ \hline
\hline

    PackNet-SfM \cite{guizilini20203d} &M &1 & 640 x 192 & 0.111 & 0.785 & 4.601 & 0.189 & 0.878 & 0.960 & 0.982 \\
    HR-Depth \cite{lyu2021hr}  &MS &1 &640 × 192 &0.107 &0.785 &4.612 &0.185 &0.887 &0.962 &0.982 \\
    Johnston \textit{et al.} \cite{johnston2020self} &M & 1 & 640 x 192 & 0.106 & 0.861 & 4.699 & 0.185 & 0.889 & 0.962 & 0.982 \\
    Monodepth2 (\textcolor{blue}{34M}) \cite{godard2019digging}  &MS &1 &640 × 192 &0.106 &0.818 &4.750 &0.196 &0.874 &0.957 &0.979 \\
    Wang \textit{et al.} \cite{wang2020self} &M &2(-1, 0) & 640 x 192 & 0.106 & 0.799 & 4.662 & 0.187 & 0.889 & 0.961 & 0.982 \\
    CADepth-Net \cite{yan2021channel} &M &1 &640 × 192 &0.105 &0.769 &4.535 &0.181 &0.892 &0.964 &\underline{0.983} \\
    DynamicDepth \cite{feng2022disentangling} &M &2(-1, 0) & 640 x 192 & 0.096 & 0.720 & 4.458 & 0.175 & 0.897 & 0.964 & \textbf{0.984} \\
    ManyDepth (MR, \textcolor{blue}{36M}) \cite{watson2021temporal} &M &2(-1, 0)+TTR & 640 x 192 & \underline{0.090} & \underline{0.713} & 4.261 & 0.170 & \underline{0.914} & 0.966 & \underline{0.983} \\
    \rowcolor{gray!25}{\textbf{SQLdepth (Efficient-b5, \textcolor{blue}{34M})}} &M &1 &640 x 192 &   0.094  &   \textbf{0.697}  &   4.320  &   0.172  &   0.904  &   0.967  &   \textbf{0.984} \\
    \rowcolor{gray!25}{\textbf{SQLdepth (ResNet-50, \textcolor{blue}{31M})}} &M &1 &640 x 192  &   0.091  &   \underline{0.713}  &   \underline{4.204}  &   \underline{0.169}  &   \underline{0.914}  &   \underline{0.968}  &   \textbf{0.984}  \\
    \rowcolor{gray!25}{\textbf{SQLdepth (ResNet-50, \textcolor{blue}{31M})}} &MS &1 &640 x 192 &   \textbf{0.088}  &   \textbf{0.697}  &   \textbf{4.175}  &   \textbf{0.167}  &   \textbf{0.919}  &   \textbf{0.969}  &   \textbf{0.984} \\
    \hline
    Monodepth2 (\textcolor{blue}{34M}) \cite{godard2019digging}  &MS &1 &1024 × 320 &0.106 &0.806 &4.630 &0.193 &0.876 &0.958 &0.980 \\
    Wang \textit{et al.} \cite{wang2020self}&M &2(-1, 0) &1024 x 320 &0.106 &0.773 &4.491 &0.185 &0.890 &0.962 &0.982 \\
    HR-Depth \cite{lyu2021hr}  &MS &1 &1024 × 320 &0.101 &0.716 &4.395 &0.179 &0.899 &0.966 &0.983 \\
    FeatDepth-MS \cite{shu2020feature} &MS &1 &1024 x 320 & 0.099		& 0.697 & 4.427 	& 0.184 & 0.889	& 0.963	& 0.982 \\
    DIFFNet \cite{zhou2021self} &M &1 &1024 x 320 & 0.097 & 0.722 & 4.345 & 0.174 & 0.907 & 0.967 & \underline{0.984} \\
    Depth Hints \cite{watson2019self} &S+Aux &1 &1024 x 320            &0.096 &0.710 &4.393 &0.185 &0.890 &0.962 &0.981   \\
    CADepth-Net \cite{yan2021channel} &MS &1 &1024 × 320 &0.096 &0.694 &4.264 &0.173 &0.908 &0.968 &\underline{0.984} \\
    EPCDepth (ResNet-50) \cite{peng2021excavating} &S+Distill &1 &1024 x 320  & 0.091 & \underline{0.646}	& 4.207 & 0.176 & 0.901	& 0.966	& 0.983 \\
    ManyDepth (ResNet-50, \textcolor{blue}{37M}) \cite{watson2021temporal} &M &2(-1, 0)+TTR &1024 x 320 & \underline{0.087} & 0.685 & 4.142 & 0.167 & \underline{0.920} & 0.968 & 0.983 \\
    \rowcolor{gray!25}\textbf{SQLdepth  (Efficient-b5, \textcolor{blue}{37M})} &M &1 &1024 x 320 &   \underline{0.087}  &   0.649  &   4.149  &   \underline{0.165}  &   0.918  &   0.969  &   \underline{0.984} \\ 
    \rowcolor{gray!25}\textbf{SQLdepth  (ResNet-50, \textcolor{blue}{37M})} &M &1 &1024 x 320 &   \underline{0.087}  &   0.659  &   \underline{4.096}  &   \underline{0.165}  &   \underline{0.920}  &   \underline{0.970}  &   \underline{0.984} \\
    \rowcolor{gray!25}\textbf{SQLdepth (ResNet-50, \textcolor{blue}{37M})} &MS &1 &1024 x 320 &   \textbf{0.082}  &   \textbf{0.607}  &   \textbf{3.914}  &   \textbf{0.160}  &   \textbf{0.928}  &   \textbf{0.972}  &   \textbf{0.985} \\
    \hline
  \end{tabular}}
  \caption{
    \textbf{Performance comparison on KITTI \cite{Geiger2013IJRR} eigen benchmark.}
    In the \textit{Train} column, \textbf{S}: trained with synchronized stereo pairs, \textbf{M}: trained with monocular videos,
    \textbf{MS}: trained with monocular videos and stereo pairs,
    \textbf{Distill}: self-distillation training, \textbf{Aux}: using auxiliary information.
    In the \textit{Test} column, \textbf{1}: one single frame as input, \textbf{2(-1, 0)}: two frames (the previous and current) as input.
    The best results are in \textbf{bold}, and second best are \underline{underlined}.
    All self-supervised methods use median-scaling in \cite{eigen2015predicting} to estimate the absolute depth scale.
  }
  \label{tab:performance}
\end{table*}

\begin{table*}[t]
\centering
  \resizebox{\textwidth}{!}{ 
  \begin{tabular}{|c|c|c|c||c|c|c|c|c|c|c|}
\hline
  Method &Train &Test &HxW &\cellcolor{red!25}$AbsRel\downarrow$ &\cellcolor{red!25}$SqRel\downarrow$ &\cellcolor{red!25}$RMSE\downarrow$ &\cellcolor{red!25}$RMESlog\downarrow$ &\cellcolor{blue!25}$\delta<1.25\uparrow$ &\cellcolor{blue!25}$\delta<1.25^2\uparrow$ &\cellcolor{blue!25}$\delta<1.25^3\uparrow$   \\ \hline
\hline
    Johnston \textit{et al.} \cite{johnston2020self} &M & 1 & 640 x 192 & 0.081 & 0.484 & 3.716 & 0.126 & 0.927 & 0.985 & 0.996 \\
    PackNet-SfM \cite{guizilini20203d} &M &1 & 640 x 192 & 0.078 & 0.420 & 3.485 & 0.121 & 0.931 & 0.986 & 0.996 \\
    Monodepth2 (\textcolor{blue}{34M}) \cite{godard2019digging}  &MS &1 &640 × 192 & 0.080 & 0.466 & 3.681 & 0.127 & 0.926 & 0.985 & 0.995 \\
    Wang \textit{et al.} \cite{wang2020self} &M &2(-1, 0) & 640 x 192 & 0.082 & 0.462 & 3.739 & 0.127 & 0.923 & 0.984 & 0.996 \\
    CADepth-Net \cite{yan2021channel}  &M &1 & 640 × 192 & 0.080 & 0.442 & 3.639 & 0.124 & 0.927 & 0.986 & 0.996 \\
    DynamicDepth \cite{feng2022disentangling} &M  &2(-1, 0) & 640 x 192 & 0.068 & 0.362 & 3.454 & 0.111 & 0.943 & 0.991 & \textbf{0.998} \\
    ManyDepth (MR, \textcolor{blue}{36M}) \cite{watson2021temporal} &M & 2(-1, 0)+TTR & 640 x 192 & \underline{0.058} & 0.334 & 3.137 & 0.101 & \underline{0.958} & 0.991 & \underline{0.997} \\
    \rowcolor{gray!25}\textbf{SQLdepth  (Efficient-b5, \textcolor{blue}{34M})} &M &1 &640 x 192 &   0.066  &   0.356  &   3.344  &   0.107  &   0.947  &   0.989  &   \underline{0.997} \\
    \rowcolor{gray!25}{\textbf{SQLdepth (ResNet-50, \textcolor{blue}{31M})}} &M &1 &640 x 192 &   0.061  &   \underline{0.317}  &   \underline{3.055}  &   \underline{0.100}  &   0.957  &   \underline{0.992}  &   \underline{0.997} \\
    \rowcolor{gray!25}{\textbf{SQLdepth (ResNet-50, \textcolor{blue}{31M})}} &MS &1 &640 x 192 &   \textbf{0.054}  &   \textbf{0.276}  &   \textbf{2.819}  &   \textbf{0.092}  &   \textbf{0.964}  &   \textbf{0.993}  &   \textbf{0.998} \\
    \hline 
    Monodepth2 (\textcolor{blue}{34M}) \cite{godard2019digging}  &MS &1 &1024 × 320 & 0.091 & 0.531 & 3.742 & 0.135 & 0.916 & 0.984 & 0.995 \\
    CADepth-Net \cite{yan2021channel}  & MS &1 & 1024 × 320 & 0.070 & 0.346 & 3.168 & 0.109 & 0.945 & 0.991 & \underline{0.997} \\
    ManyDepth (ResNet-50, \textcolor{blue}{37M}) \cite{watson2021temporal} &M &2(-1, 0) + TTR & 1024 x 320 &\underline{0.055} &0.305 &2.945 &\underline{0.094} &\underline{0.963} &0.992 &\underline{0.997} \\
    \rowcolor{gray!25}\textbf{SQLdepth  (Efficient-b5, \textcolor{blue}{37M})} &M &1 &1024 x 320 &   0.058  &   \underline{0.287}  &   3.039  &   0.096  &   0.959  &   0.992  &   \textbf{0.998} \\
    \rowcolor{gray!25}\textbf{SQLdepth  (ResNet-50, \textcolor{blue}{37M})} &M &1 &1024 x 320 &   0.058  &   0.289  &   \underline{2.925}  &   0.095  &   0.962  &   \underline{0.993}  &   \textbf{0.998} \\
    \rowcolor{gray!25}\textbf{SQLdepth (ResNet-50, \textcolor{blue}{37M})} &MS &1 &1024 x 320 &   \textbf{0.052}  &   \textbf{0.223}  &   \textbf{2.550}  &   \textbf{0.084}  &   \textbf{0.971}  &   \textbf{0.995}  &   \textbf{0.998} \\
    \hline
  \end{tabular}}
  \caption{
    \textbf{Performance comparison using KITTI improved ground truth from  \cite{uhrig2017sparsity}.
    } 
TTR is the test-time refinement used by ManyDepth.
  }
  \label{tab:improved performance}
\end{table*}

\begin{table*}[t]
\centering
  \resizebox{\textwidth}{!}{ 
  \begin{tabular}{|c|c|c|c||c|c|c|c|c|c|c|}
\hline
  Method &Train &Test &HxW &\cellcolor{red!25}$AbsRel\downarrow$ &\cellcolor{red!25}$SqRel\downarrow$ &\cellcolor{red!25}$RMSE\downarrow$ &\cellcolor{red!25}$RMESlog\downarrow$ &\cellcolor{blue!25}$\delta<1.25\uparrow$ &\cellcolor{blue!25}$\delta<1.25^2\uparrow$ &\cellcolor{blue!25}$\delta<1.25^3\uparrow$   \\ \hline
\hline
    Pilzer \textit{et al.} \cite{AndreaPilzer2018UnsupervisedAD} &GAN, C & 1 & 512 x 256 & 0.240 & 4.264 & 8.049 & 0.334 & 0.710 & 0.871 & 0.937 \\
    Struct2Depth 2 \cite{VincentCasser2019UnsupervisedMD}  &MMask, C & 1 & 416 x 128 & 0.145 & 1.737 & 7.280 & 0.205 & 0.813 & 0.942 & 0.976 \\
    Monodepth2 \cite{godard2019digging}  &–, C & 1 & 416 x 128 & 0.129 & 1.569 & 6.876 & 0.187 & 0.849 & 0.957 & 0.983 \\
    Videos in the Wild \cite{gordon2019depth}  &MMask, C & 1 & 416 x 128 & 0.127 & 1.330 & 6.960 & 0.195 & 0.830 & 0.947 & 0.981 \\
    \rowcolor{pink!25} {\textbf{SQLdepth (Zero-shot)}} &–, K & 1 &416 x 128 &0.125 &1.347 &7.398 &0.194 &0.834 &0.951 &0.985 \\
    Li \textit{et al.} \cite{HanhanLi2020UnsupervisedMD} &MMask, C & 1 & 416 x 128 & 0.119 & 1.290 & 6.980 & 0.190 & 0.846 & 0.952 & 0.982 \\
    Lee \textit{et al.} \cite{SeokjuLee2021AttentiveAC}  &MMask, C & 1 & 832 x 256 & 0.116 & 1.213 & 6.695 & 0.186 & 0.852 & 0.951 & 0.982 \\
    ManyDepth \cite{watson2021temporal}  &MMask, C & 2(-1, 0) & 416 x 128 & 0.114 & 1.193 & \textbf{6.223} & 0.170 & 0.875 & 0.967 & 0.989 \\
    InstaDM \cite{SeokjuLee2021LearningMD}  &MMask, C & 1 & 832 x 256 & 0.111 & \underline{1.158} & 6.437 & 0.182 & 0.868 & 0.961 & 0.983 \\
    \rowcolor{green!25}{\textbf{SQLdepth (From scratch)}} &–, C &1 &416 x 128 & \underline{0.110}  &\textbf{1.130}  &6.264  &\underline{0.165}  &\underline{0.881}  &\underline{0.971}  &\textbf{0.991} \\
    \rowcolor{gray!25}{\textbf{SQLdepth (Fine-tuned)}} &–, K$\rightarrow$C &1 &416 x 128    &   \textbf{0.106}  &   1.173  &   \underline{6.237}  &   \textbf{0.163}  &   \textbf{0.888}  &   \textbf{0.972}  &   \underline{0.990} \\
    \hline 
  \end{tabular}}
  \caption{
    \textbf{Performance comparison on Cityscapes \cite{cordts2016cityscapes} dataset. }
    We present results of zero-shot, fine-tuning (self-supervised) 
    and training from scratch on Cityscapes.
    All the other baselines are trained from scratch on Cityscapes.
    \textbf{K} for KITTI, \textbf{C} for Cityscapes, \textbf{K$\rightarrow$C} for pretrained on KITTI and fine-tuned on Cityscapes.
    \textbf{MMask} means using motion mask to deal with moving objects (\textbf{extremely important for training on Cityscapes}),
    \textbf{–} for no motion mask.
  }
  \label{tab:performance cityscapes}
\end{table*}

\begin{figure}[htbp]
  \setlength{\columnsep}{3pt} 
\begin{multicols}{2}
  \centering

  \begin{subfigure}[b]{0.24\textwidth}
    \begin{minipage}[t]{\textwidth}
      \centering
      \includegraphics[width=\textwidth]{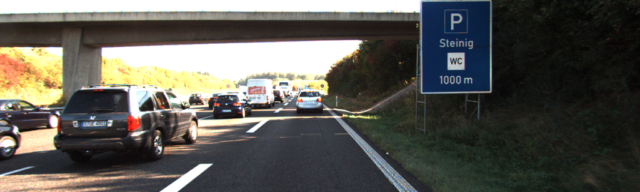}
      \includegraphics[width=\textwidth]{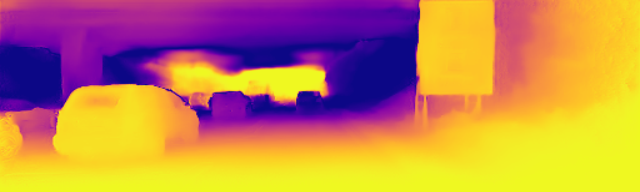}
      \includegraphics[width=\textwidth]{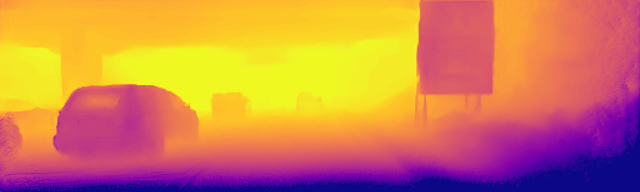}
    \end{minipage}%
  \end{subfigure}

  \begin{subfigure}[b]{0.24\textwidth}
    \begin{minipage}[t]{\textwidth}
      \centering
      \includegraphics[width=\textwidth]{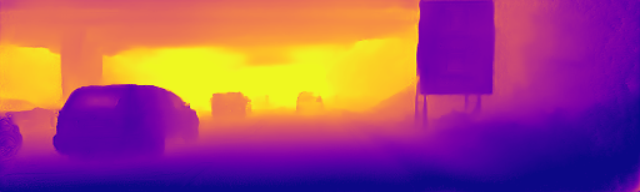}
      \includegraphics[width=\textwidth]{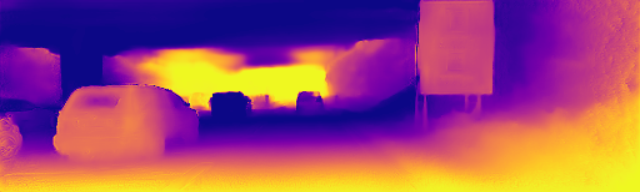}
      \includegraphics[width=\textwidth]{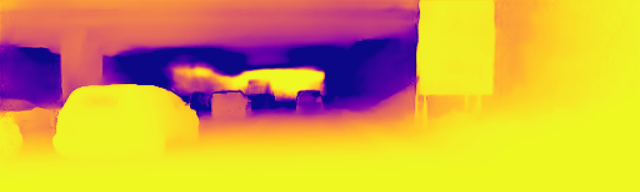}
    \end{minipage}%
  \end{subfigure}

\end{multicols}
  \centering
  \caption{\textbf{Visualization of planes in the self-cost volume.}
  It's noteworthy that each slice of the self-cost volume maintains clear scene structures.
  In contrast, feature maps do not exhibit any scene structures in general.
  This demonstrates that the self-cost volume essentially serves as a beneficial inductive bias
  that captures useful geometrical cues for depth estimation.
  }
  \label{fig:cost}
\end{figure}

\begin{figure*}[htbp]
\begin{multicols}{4}
  \centering

  \begin{subfigure}[b]{0.26\textwidth}
    \begin{minipage}[t]{\textwidth}
      \centering
      \caption{RGB}
      \includegraphics[width=\textwidth]{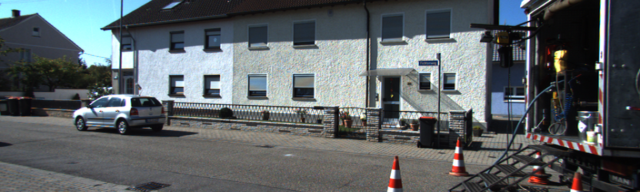}
      \includegraphics[width=\textwidth]{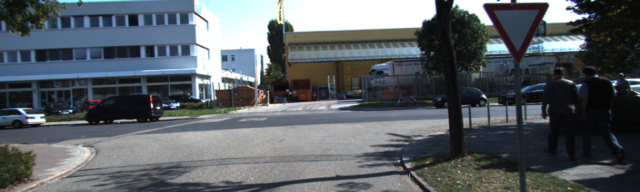}
      \includegraphics[width=\textwidth]{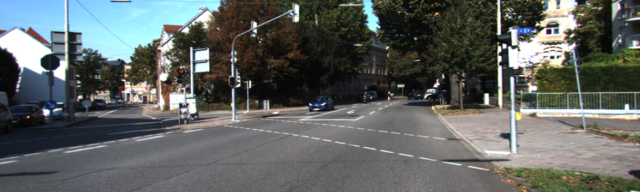}
    \end{minipage}%
  \end{subfigure}

  \begin{subfigure}[b]{0.26\textwidth}
    \begin{minipage}[t]{\textwidth}
      \centering
      \caption{Monodepth2(MS) \cite{godard2019digging}}
      \includegraphics[width=\textwidth]{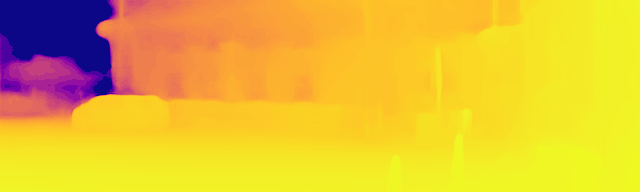}
      \includegraphics[width=\textwidth]{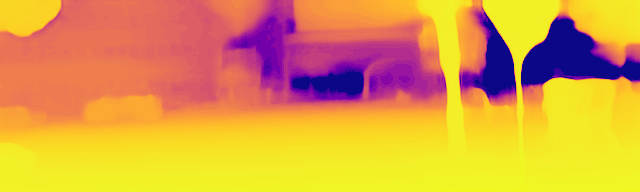}
      \includegraphics[width=\textwidth]{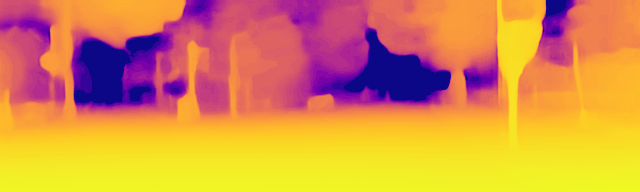}
    \end{minipage}%
  \end{subfigure}

  \begin{subfigure}[b]{0.26\textwidth}
    \begin{minipage}[t]{\textwidth}
      \centering
      \caption{EPCDepth \cite{peng2021excavating}}
      \includegraphics[width=\textwidth]{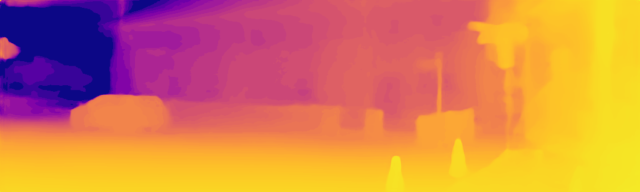}
      \includegraphics[width=\textwidth]{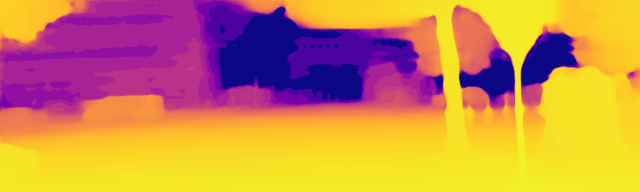}
      \includegraphics[width=\textwidth]{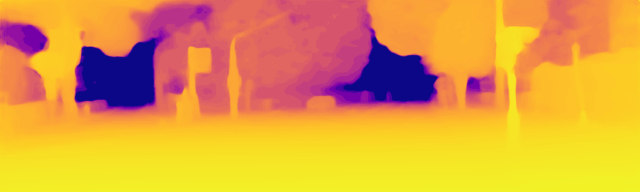}
    \end{minipage}%
  \end{subfigure}

  \begin{subfigure}[b]{0.26\textwidth}
    \begin{minipage}[t]{\textwidth}
      \centering
      \caption{SQLdepth (Ours)}
      \includegraphics[width=\textwidth]{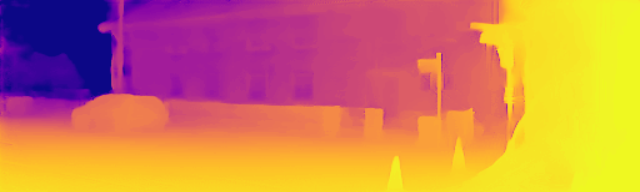}
      \includegraphics[width=\textwidth]{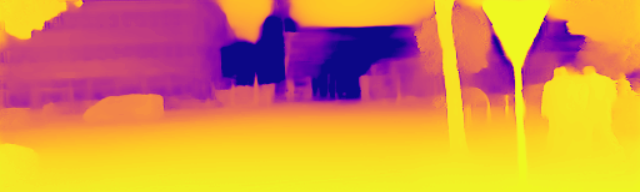}
      \includegraphics[width=\textwidth]{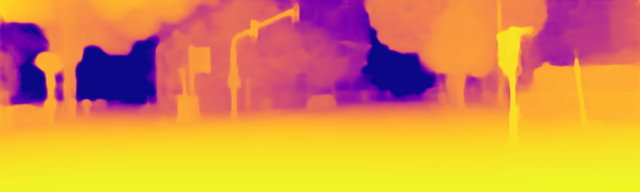}

    \end{minipage}%
  \end{subfigure}

\end{multicols}
  \centering
  \caption{\textbf{Additional Qualitative results on the KITTI eigen benchmark.}}
  \label{fig:kitti_eval}
\end{figure*}

\subsection{Loss Function}
\label{sec:pe}

\noindent\textbf{Objective Functions.}
Following  \cite{godard2017unsupervised, godard2019digging}, we use the standard photometric loss $pe$
combined by the L1 and SSIM \cite{wang2004image} as Eq. \ref{loss}.
\begin{equation}
  \label{loss}
p e\left(I_a, I_b\right)=\frac{\alpha}{2}\left(1-\operatorname{SSIM}\left(I_a,
I_b\right)\right)+(1-\alpha)\left\|I_a-I_b\right\|_1
\end{equation}
In order to regularize the depths in
textureless regions, we use edge-aware smooth loss as follows:
\begin{equation}
  \label{smooth loss}
L_s=\left|\partial_x d_t^*\right|
e^{-\left|\partial_x I_t\right|}+\left|\partial_y d_t^*\right|
e^{-\left|\partial_y I_t\right|} 
\end{equation}
\noindent\textbf{Masking Strategy.}
In real-world scenarios, situations like stationary camera and moving objects
will break down the assumptions of a moving camera and a static scene 
and hurt the performance of self-supervised depth estimator. To improve the
accuracy of depth prediction, several prior studies integrated motion mask to
deal with moving objects with the help of a scene specific instance
segmentation model (especially in Cityscapes), but at the same time, their
methods are thus unable to expand to unseen scenarios. 
To keep scalable, we do not use motion mask to deal with moving objects.
Instead,
we just use auto-masking strategy in \cite{godard2019digging} to filter out stationary pixels and low-texture regions that remain 
with the same appearance between two frames in a sequence.
The binary mask $\mu$ is computed in Eq. \ref{mask},
where [] is the Iverson bracket.
\begin{equation}
  \label{mask}
\mu=\left[\min _{t^{\prime}} p e\left(I_t, I_{t^{\prime} \rightarrow
t}\right)<\min _{t^{\prime}} p e\left(I_t, I_{t^{\prime}}\right)\right]
\end{equation}

\noindent\textbf{Final Training Loss.}
We combine our per-pixel smooth loss and masked photometric losses as Eq. \ref{final_loss}.
\begin{equation}
\label{final_loss}
L=\mu L_p+\lambda L_s 
\end{equation}
\section{Experiments}
We evaluate the performance of SQLdepth on three public datasets including KITTI, Cityscapes and Make3D,
and quantify the model performance in terms of several widely used metrics from  \cite{eigen2015predicting}. 
In addition, we investigate the generalization of our model
by zero-shot evaluating or fine-tuning on new datasets with KITTI-pretrained weights.
\subsection{Datasets and Experimental Protocol}
\noindent\textbf{KITTI}  \cite{Geiger2013IJRR} is a dataset that provides stereo image sequences, 
which is commonly used for self-supervised monocular depth estimation. 
We use Eigen split  \cite{eigen2015predicting} with around 26k images for training and 697 for testing. 
We train SQLdepth from scratch on KITTI with the least requirements: 
\textbf{no motion mask (auto-masking \cite{godard2019digging} only), no extra stereo pairs, and no auxiliary information.}
For testing, we also keep the hardest setting: \textbf{only one single frame as input,} 
while the other methods may use multiple frames as input to improve accuracy.

\noindent\textbf{Cityscapes} \cite{cordts2016cityscapes} is a challenging dataset which contains numerous moving objects. 
In order to evaluate the generalization of SQLdepth, we fine tune and perform zero-shot evaluation on Cityscapes using the KITTI pre-trained model respectively.
We have to emphasize that 
\textbf{we do not use motion mask while most of the other baselines do.}
We use the data preprocessing scripts from \cite{2017Unsupervised} as others baselines do,
and preprocess the image sequences into triples.
In addition, we also train SQLdepth from scratch on Cityscapes under the same setting with other baselines. 

\noindent\textbf{Make3D} \cite{saxena2008make3d} To evaluate the generalization ability of SQLdepth on unseen images, we use the KITTI-pretrained SQLdepth to perform zero-shot evaluation on the Make3D dataset,
and provide additional depth map visualizations.

\subsection{Implementation Details}\label{sec:details}
Our model is implemented with Pytorch framework \cite{AdamPaszke2019PyTorchAI}. 
The model is trained on 4 NVIDIA GTX 2080Ti GPUs, with a batch size of 8.
Following the settings from \cite{godard2019digging}, we use color and flip augmentations on images during training. 
We jointly train both DepthNet and PoseNet with the
Adam Optimizer \cite{DiederikPKingma2014AdamAM} with $\beta_1 = 0.9$, $\beta_2 = 0.999$. 
The initial learning rate is set to $1e-4$ and decays to $1e-5$ after $15$ epochs.
We set the SSIM weight to $\alpha = 0.85$ and smooth loss term weight to $\lambda = 1e-3$.
We use the ResNet-50 \cite{he2016deep} with ImageNet \cite{russakovsky2015imagenet} pretrained weights as backbone, 
as the other baselines do.
We also provide results of ImageNet pretrained efficient-net-b5
\cite{MingxingTan2019EfficientNetRM} backbone, which has similar params with
ResNet-50.
\subsection{KITTI Results}
\begin{figure}[t]
 \centering
  \centering
  \tiny
      \begin{overpic}[width=\linewidth,keepaspectratio]{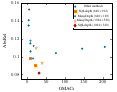}
        \put(29.5,28){\cite{watson2021temporal}}
        \put(24.5,45){\cite{watson2021temporal}}
        \put(35,26){\cite{watson2021temporal}}
        \put(31.5,36.5){\cite{watson2021temporal}}
        \put(23.5,72){\cite{poggi2018towards}}
        \put(22.5,63){\cite{casser2019depth}}
        \put(19.5,57.5){\cite{YuhuaChen2019SelfSupervisedLW}}
        \put(21.5,33){\cite{wang2020self}}
        \put(46.5,34){\cite{shu2020feature}}
        \put(65.5,38){\cite{guizilini20203d}}
        \put(26.5,40){\cite{godard2019digging}}
        \put(78.5,40){\cite{luo2020consistent}}

      \end{overpic}

   \caption{
     \textbf{Comparisons about computational costs.}
     We compare AbsRel against Giga Multiply-Add Caculation per Second (GMACs) on the KITTI Eigen test set.
     Our model is more efficient and accurate, especially for the ResNet-50 based 1024$\times$320 model 
     (closest to the $(0, 0)$ point).
   }
   \label{img:MACs}
\end{figure}
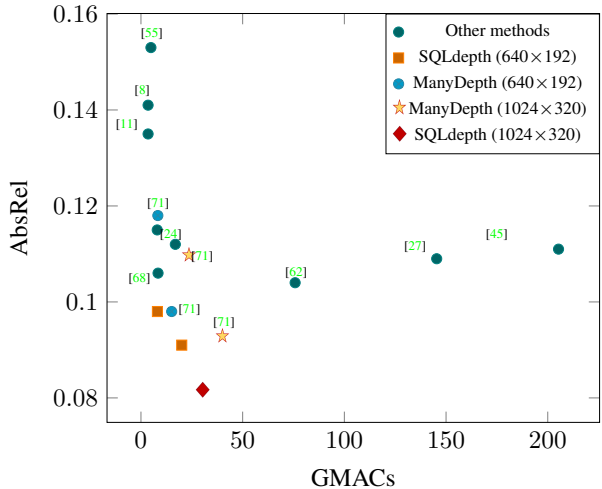

The results are presented in Table \ref{tab:performance}.
Following previous works, we conduct experiments under two different resolutions.
In the top half, 
the resolution of input image is 640$\times$192,
and the bottom half for high resolution of 1024$\times$320. 
We observe that SQLdepth outperforms all existing 
self-supervised methods by significant margins,
and also outperforms counterparts trained with additional stereo pairs, or use multi-frames for testing.
Furthermore, we conducted a comparative study with Monodepth2 and EPCDepth. 
As shown in Figure \ref{fig:kitti_eval} and Figure \ref{img:first}, 
SQLdepth produces impressive depth maps with sharp boundaries,
especially for fine-grained scene details, such as traffic signs and pedestrians. 
Regarding for why our method can effectively recover scene details, the possible
reasons are analyzed in Figure \ref{fig:cost}.
As for efficiency comparison, details are present in Figure \ref{img:MACs}.

Due to the low quality of ground truth in KITTI, we also provide evaluation
results with KITTI improved ground-truth in Table \ref{tab:improved
performance}. Compared with ManyDepth \cite{watson2021temporal}, which 
uses multiple frames for testing, SQLdepth still presents better results across
all metrics, and achieves a $6.5\%$ error reduction in terms of AbsRel in
1024$\times$320 resolution, and $12.85\%$ error reduction in 640$\times$192
resolution.

\subsection{Cityscapes Results}
In order to evaluate the generalization of SQLdepth, 
we present results of zero-shot evaluation, fine-tuning (self-supervised and without motion mask),
and training from scratch on Cityscapes.
We used the KITTI pre-trained model for zero-shot evaluation and fine-tuning. 
The results are reported in Table \ref{tab:performance cityscapes}. 
We have to emphasize that
although most of the baselines in Table \ref{tab:performance cityscapes} use motion mask to deal with moving objects,
SQLdepth still presents overwhelming performance without motion mask,
and achieves a $4.4\%$ error reduction from the previous best method InstaDM \cite{SeokjuLee2021LearningMD}.
Moreover, we noticed that SQLdepth converges fast and only takes 2 epochs to achieve the rank 1$^{st}$.
The pink row of zero-shot evaluation in Table \ref{tab:performance cityscapes} shows competitive performance.
In addition, the green row of training from scratch also shows state-of-the-art performance.
These demonstrate the superior generalization of SQLdepth.

\subsection{Make3D results}
\begin{figure}[ht]
  \setlength{\columnsep}{0.4pt} 
\begin{multicols}{4}
  \centering

  \begin{subfigure}[b]{0.1175\textwidth}
    \begin{minipage}[t]{\textwidth}
      \centering
      \caption{\footnotesize{RGB}}
      \includegraphics[width=\textwidth]{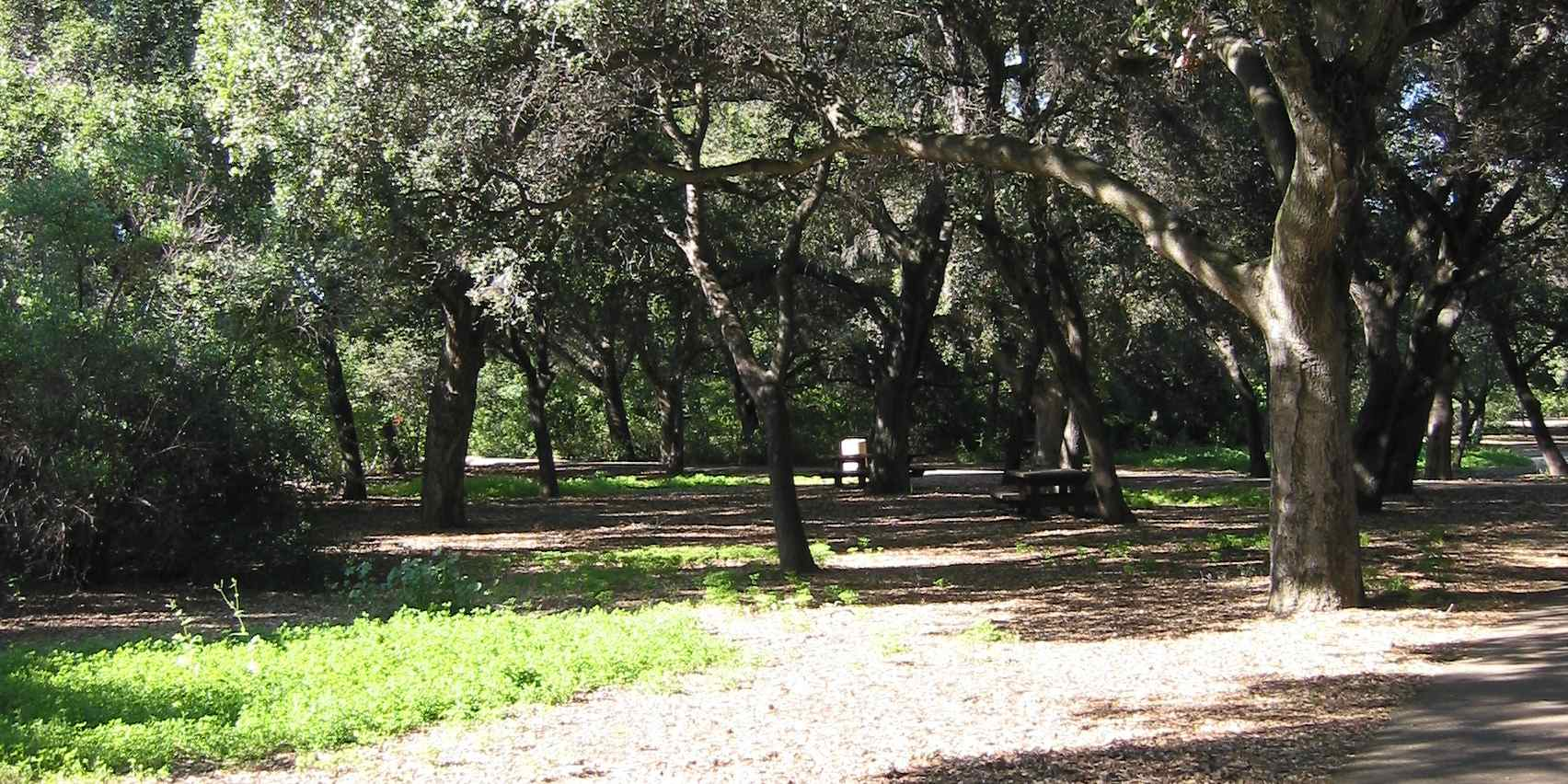}
      \includegraphics[width=\textwidth]{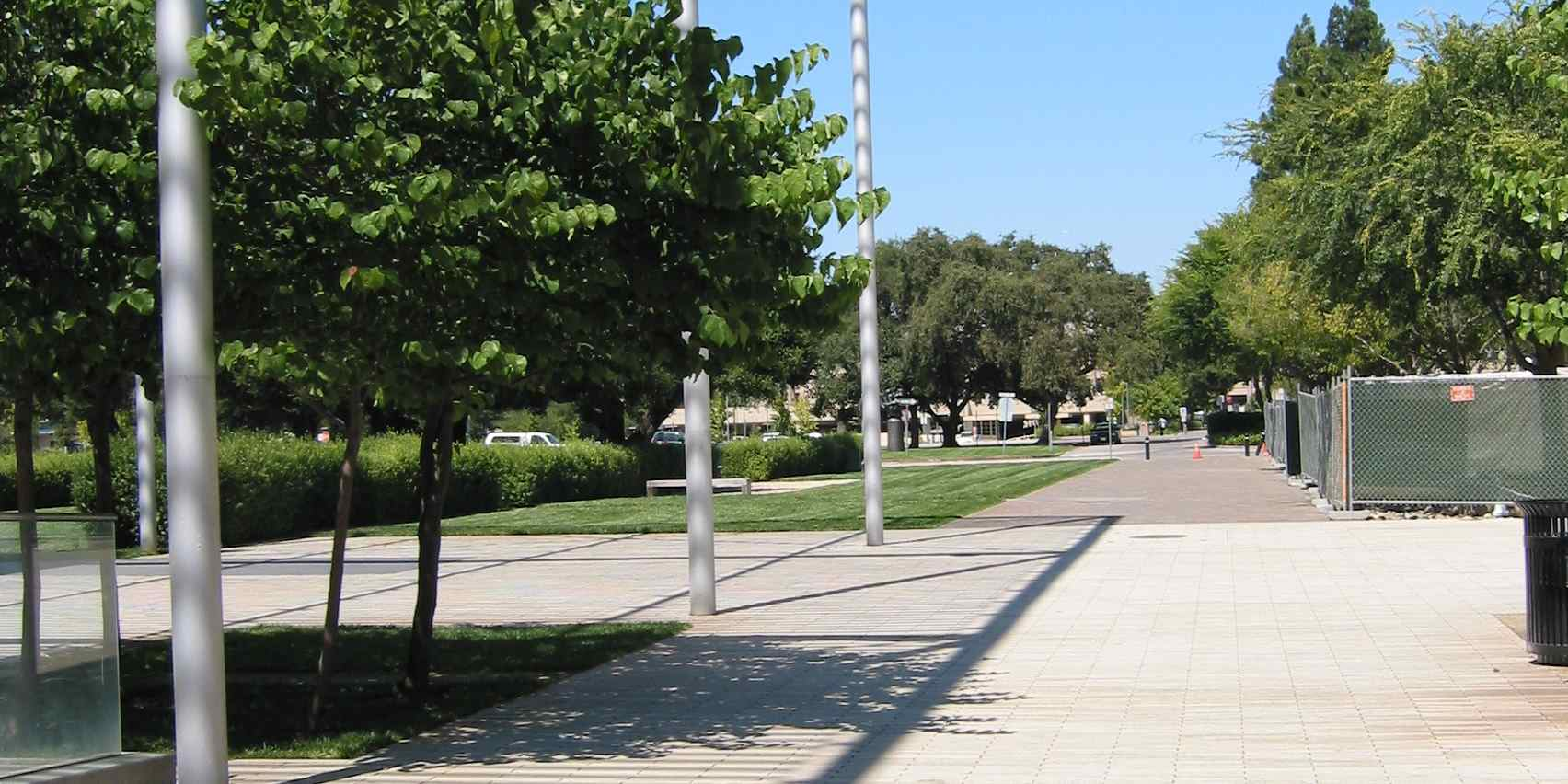}
      \includegraphics[width=\textwidth]{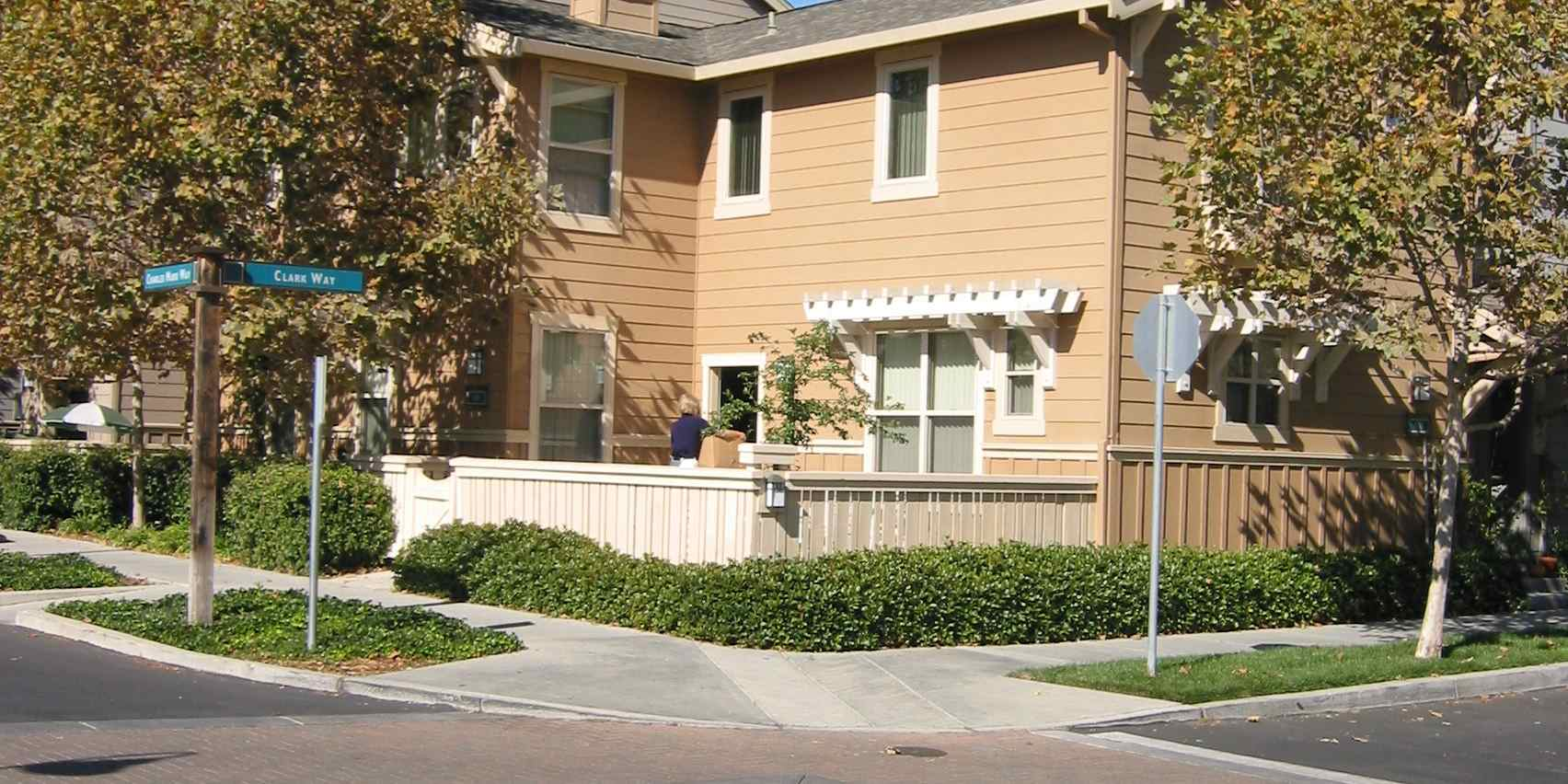}
    \end{minipage}%
  \end{subfigure}

  \begin{subfigure}[b]{0.1175\textwidth}
    \begin{minipage}[t]{\textwidth}
      \centering
      \caption{\footnotesize{MD2(M) \cite{godard2019digging}}}
      \includegraphics[width=\textwidth]{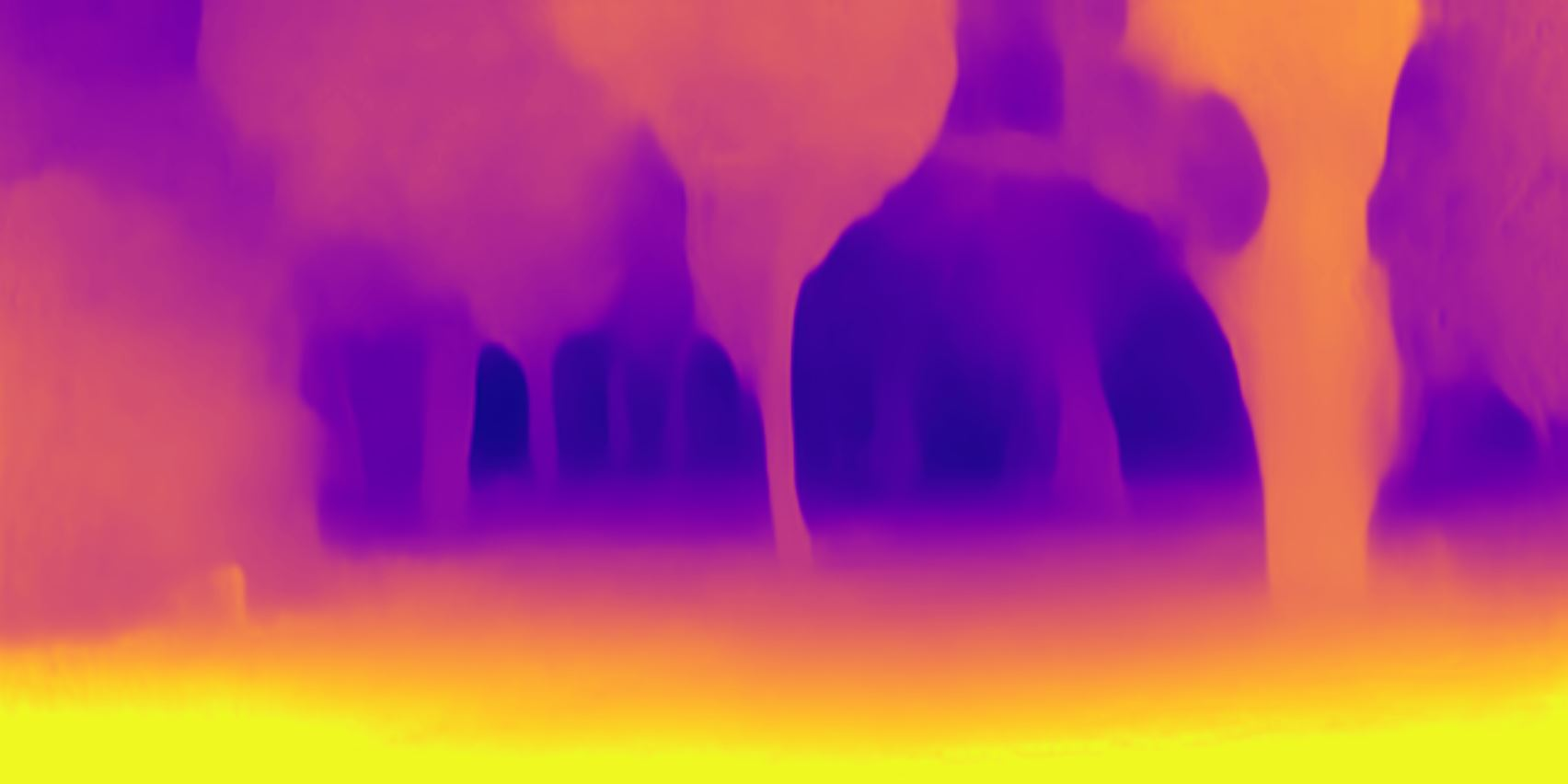}
      \includegraphics[width=\textwidth]{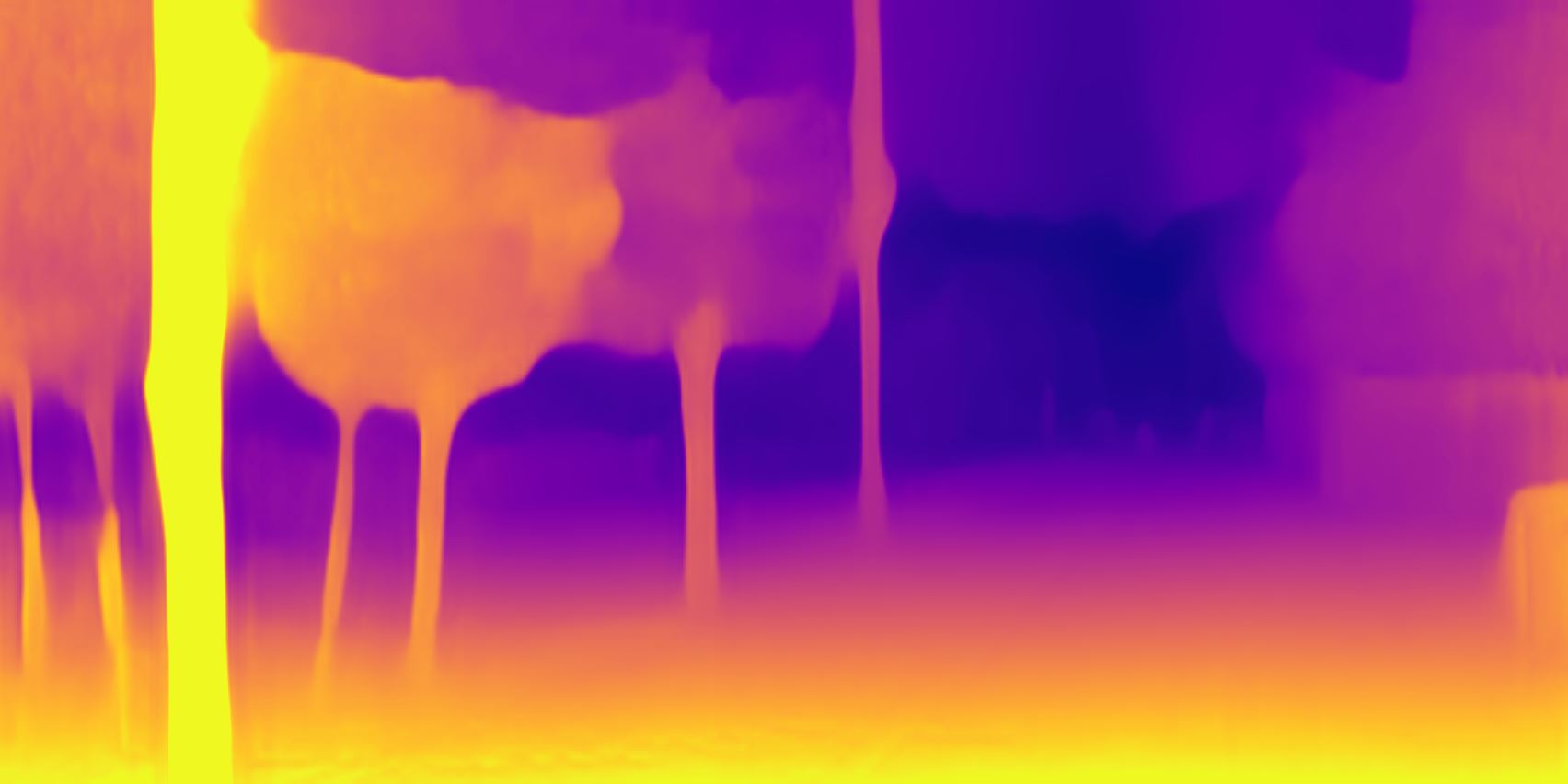}
      \includegraphics[width=\textwidth]{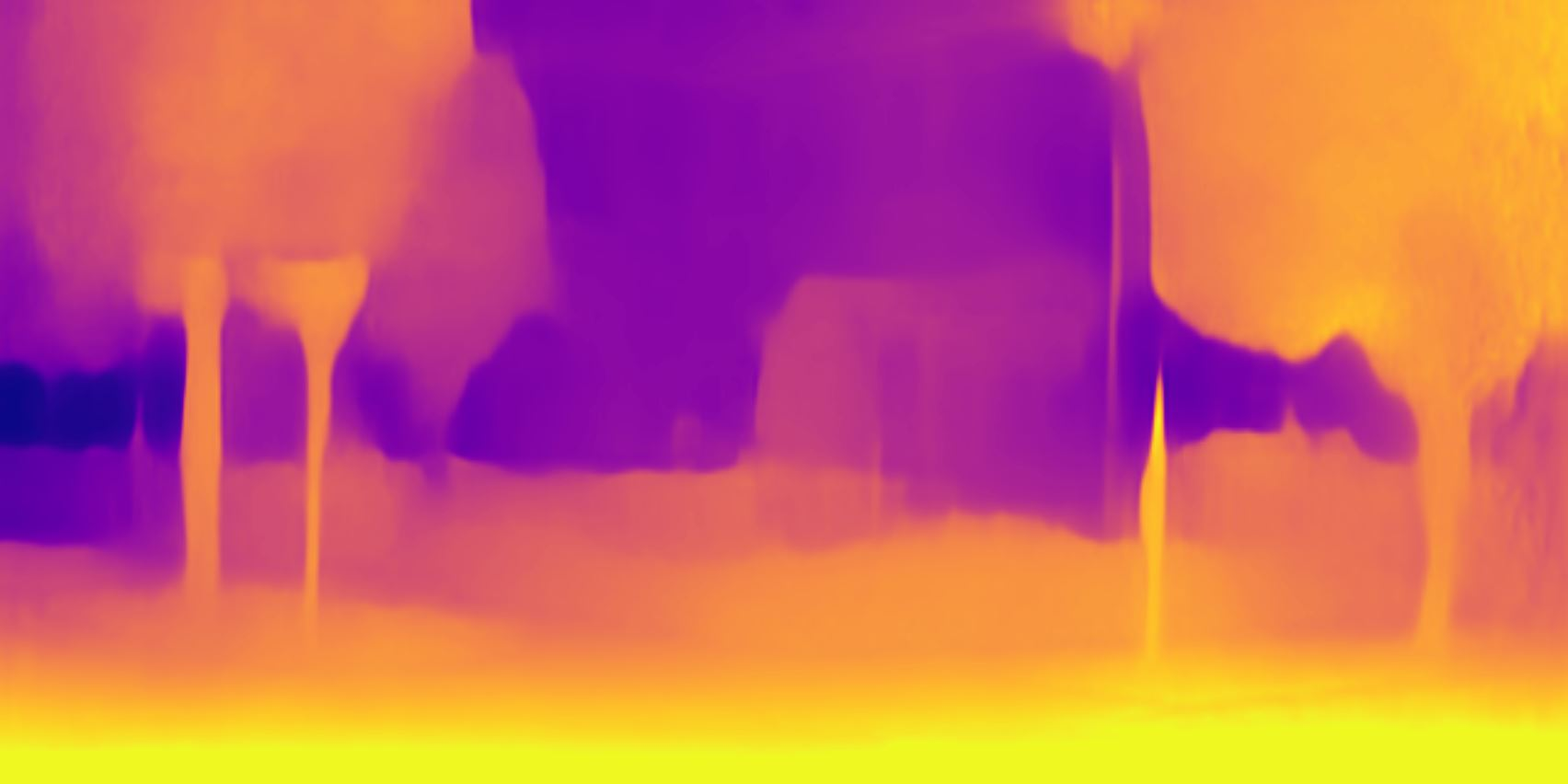}
    \end{minipage}%
  \end{subfigure}

  \begin{subfigure}[b]{0.1175\textwidth}
    \begin{minipage}[t]{\textwidth}
      \centering
      \caption{\footnotesize{Ours}}
      \includegraphics[width=\textwidth]{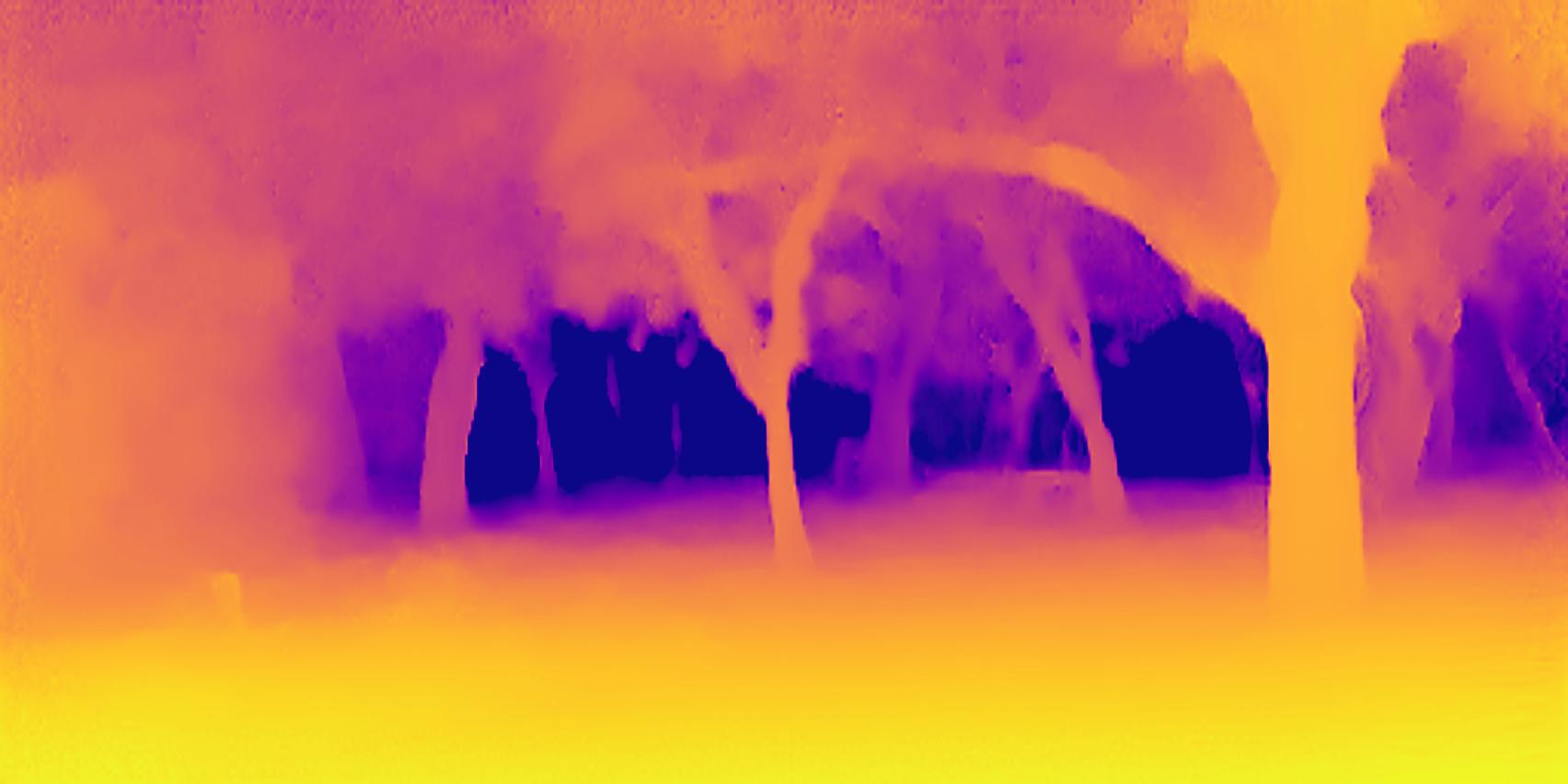}
      \includegraphics[width=\textwidth]{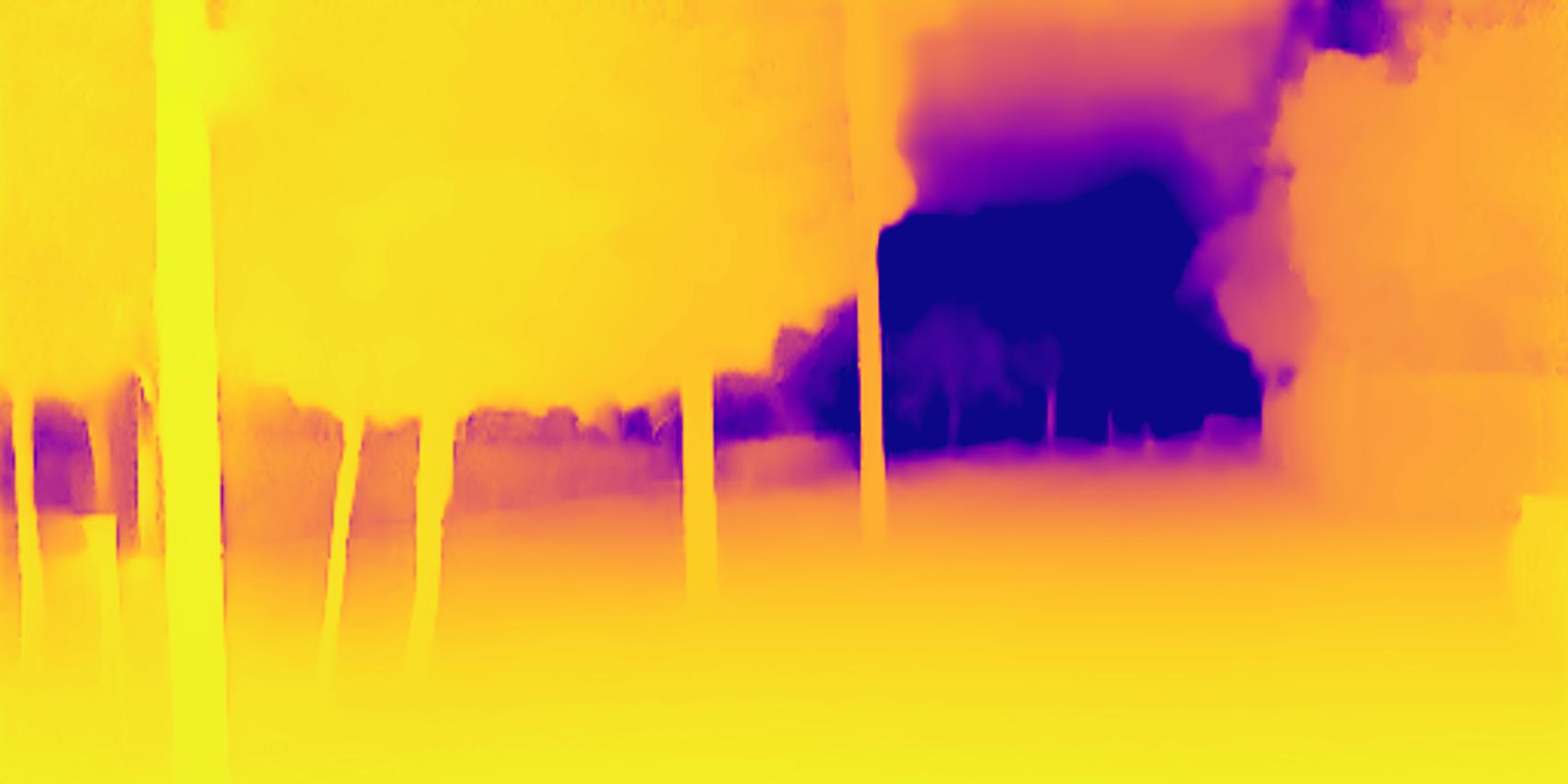}
      \includegraphics[width=\textwidth]{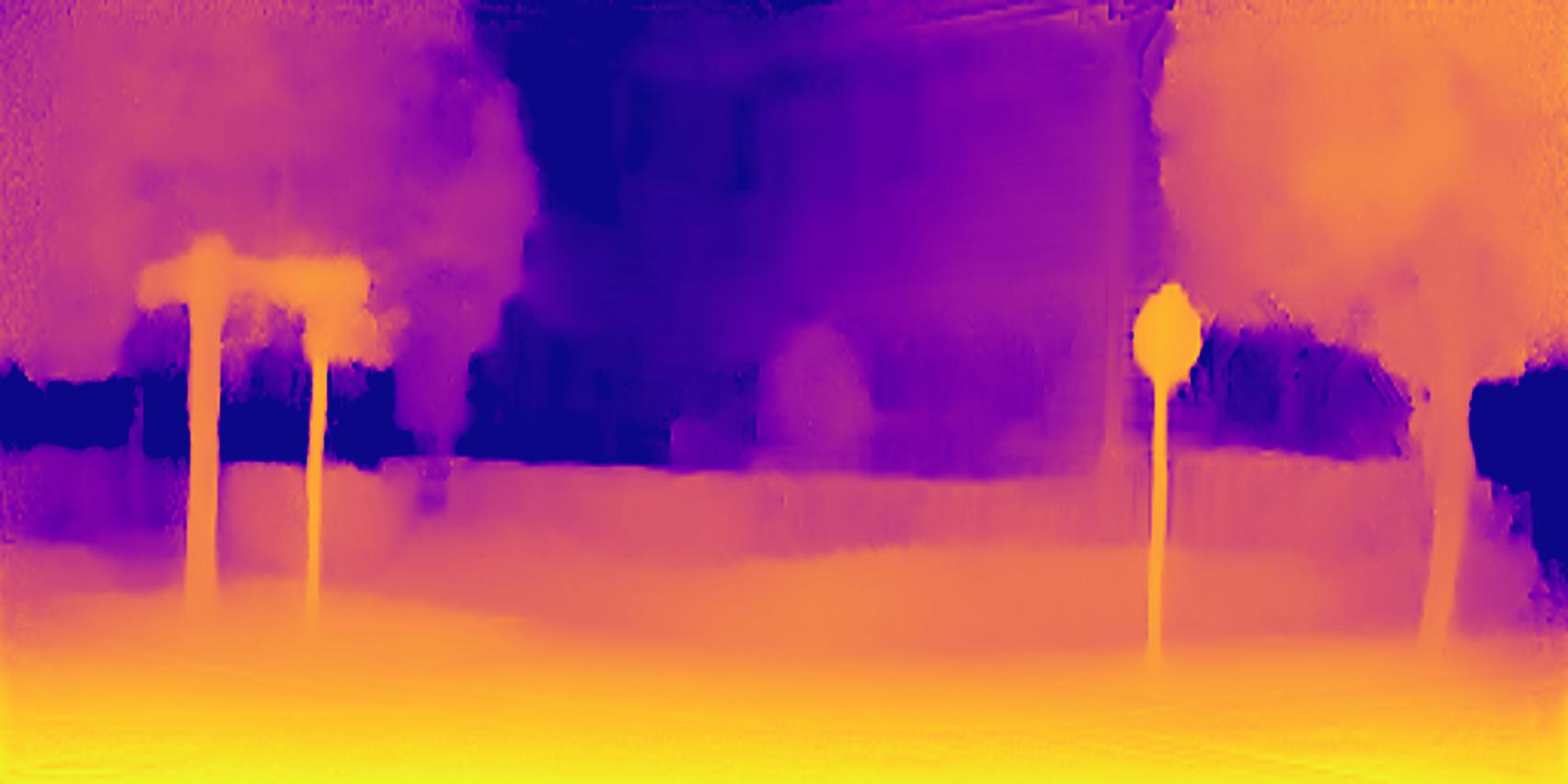}
    \end{minipage}%
  \end{subfigure}

  \begin{subfigure}[b]{0.1175\textwidth}
    \begin{minipage}[t]{\textwidth}
      \centering
      \caption{\footnotesize{gt}}
      \includegraphics[width=\textwidth]{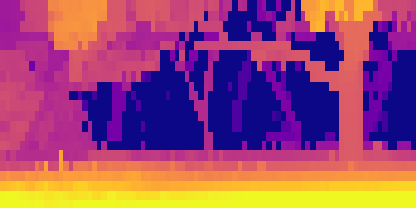}
      \includegraphics[width=\textwidth]{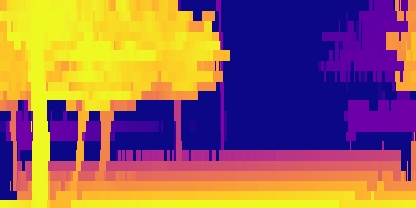}
      \includegraphics[width=\textwidth]{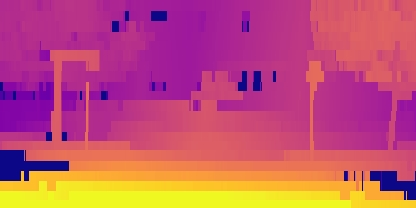}
    \end{minipage}%
  \end{subfigure}

\end{multicols}
  \centering
  \caption{\textbf{Qualitative Make3D results (Zero-shot).}
  }
  \label{fig:Make3D_eval}
\end{figure}

To further evaluate the generalization capacity of SQLdepth, we directly evaluated (zero-shot) on Make3D dataset \cite{saxena2008make3d} 
using the pretrained weights on KITTI. 
Following the same evaluation setting in \cite{godard2017unsupervised}, 
we tested on a center crop of $2 \times 1$ ratio. As shown in Table \ref{table:Make3D_metrics} and Figure \ref{fig:Make3D_eval},
SQLdepth produces superior results compared with baselines,
and produces sharp depth maps with more accurate scene details.
These demonstrate the excellent zero-shot generalization ability of our model.
\begin{table}[ht]
\centering
  \resizebox{0.48\textwidth}{!}{ 
  \begin{tabular}{|c|c||c|c|c|c|}
\hline
    Method &Type &\cellcolor{red!25}$AbsRel\downarrow$ &\cellcolor{red!25}$SqRel\downarrow$ &\cellcolor{red!25}$RMSE\downarrow$ &\cellcolor{red!25}$log_{10}\downarrow$\\ \hline
\hline
    Monodepth \cite{godard2017unsupervised}  &S &0.544 &10.94 &11.760 &0.193 \\
    Zhou \cite{zhou2017unsupervised}  &M &0.383 &5.321 &10.470 &0.478 \\
    DDVO \cite{ChaoyangWang2017LearningDF}  &M &0.387 &4.720 &8.090 &0.204 \\
    Monodepth2 \cite{godard2019digging} &M &0.322 &3.589 &7.417 &0.163 \\
    CADepthNet \cite{yan2021channel}  &M &0.312 &3.086 &7.066 &0.159 \\
    \rowcolor{gray!25}SQLdepth(Ours)  &M &\textbf{0.306} &\textbf{2.402} &\textbf{6.856} &\textbf{0.151} \\
  \hline
  \end{tabular}}
  \caption{\textbf{Make3D results.}}
  \label{table:Make3D_metrics}
\end{table}

\subsection{Ablation Study}\label{sec:ablation}
In this section, we conduct ablation studies to investigate the effects of 
designs in SQLdepth, including coarse-grained queries, plane-wise depth
counting approximation, and probabilistic combination,
respectively.

\noindent\textbf{Benefits of the SQL layer.} 
As reported in Table \ref{ablation:query}, For SQLdepth without SQL layer, 
it shows significant performance downgrade with a 17.54\% decrease from 0.091 to 0.114 in 
terms of AbsRel. 
This demonstrates the effectiveness of SQL layer.

\begin{table}[h]
\centering
  \resizebox{0.48\textwidth}{!}{ 
  \begin{tabular}{lccc}
\toprule[1pt]
    Ablation &$AbsRel\downarrow$ &$SqRel\downarrow$ &$RMSE\downarrow$ \\
\hline
    No queries (encoder-decoder only)  &0.114  &0.805     &4.816     \\      
    Learned queries (static) &0.102  &0.738     &4.512     \\      
    Fine-grained queries  &0.094  &0.727     &4.437     \\      
    \hline
    \textbf{Coarse-grained queries}  &\textbf{0.091} &\textbf{0.714} &\textbf{4.204}  \\
    \bottomrule[1pt]
  \end{tabular}}
  \caption{
    Ablation study for the core component of SQLdepth.
    Self Query Layer with runtime generated coarse-grained queries leads to better scores.
  }
  \label{ablation:query}
\end{table}
\noindent\textbf{Benefits of coarse-grained queries.} 
To investigate the effectiveness of our coarse-grained queries, we compare it with two variants of queries: learned queries (static) and fine-grained queries.
Learned queries are learnable vectors \cite{NicolasCarion2020EndtoEndOD} during training and then frozen for testing.
Fine-grained queries are generated from patch embeddings at runtime, but with smaller patch size 
(e.g. 4$\times$4). The results are summarized in Table \ref{ablation:query}.
We found that the models with runtime generated queries (fine-grained queries or
coarse-grained queries) are better than that with the static queries. 
This is due to that static queries are not able to
adaptively represent the context in different images.
For runtime queries, we noticed that coarse-grained queries
produce slightly better results compared with fine-grained queries. 
This is due to that with a larger patch size, coarse-grained queries are able to
capture the visual contexts within a larger receptive field. 
In addition, the coarse-grained queries are more computational efficient (O($h \times
w/p^2$)).

\begin{table}[h]
\centering
  \resizebox{0.48\textwidth}{!}{ 
  \begin{tabular}{lccc}
\toprule[1pt]
    Ablation &$AbsRel\downarrow$ &$SqRel\downarrow$ &$RMSE\downarrow$ \\
\hline
    Fixed bins (uniform)  &0.114  &0.894     &4.659     \\       
    Bins from direct regression \cite{ShariqFarooqBhat2020AdaBinsDE} &0.112 &0.874 &4.534 \\     
    \hline
    \textbf{Bins from depth counting approximation} &\textbf{0.087} &\textbf{0.659} &\textbf{4.096} \\     
    \bottomrule[1pt]
  \end{tabular}}
  \caption{
    Performance comparison of different design choices for depth bins.
  }
  \label{ablation:approximation}
\end{table}

\noindent\textbf{Benefits of plane-wise depth counting approximation.} 
As shown in Table \ref{ablation:approximation},
we replace depth bins 
with fixed bins or depth bins directly regressed 
from a global context extracted by a Transformer, as AdaBins \cite{ShariqFarooqBhat2020AdaBinsDE} does.
Compared with both variants, our proposed plane-wise depth counting approximation leads to better results. 
This is because that the depth counting approximation can make better use of pixel-level 
fine-grained clues.

\noindent\textbf{Benefits of probabilistic combination.} As shown in Table \ref{ablation:combination}, 
we replace the probabilistic combination with either global average pooling or
Conv1$\times$1. We observe that both of them lead to a substantial performance
drop. 
The potential reason could be that probabilistic combination 
can adaptively fuse all depth estimations provided by
different contexts in the image.

\begin{table}[h]
\centering
  \resizebox{0.48\textwidth}{!}{ 
  \begin{tabular}{lccc}
\toprule[1pt]
    Ablation &$AbsRel\downarrow$ &$SqRel\downarrow$ &$RMSE\downarrow$ \\
\hline
    Global average pooling &0.115  &0.926     &4.832     \\       
    Conv1$\times$1 as combination &0.106  &0.826 &4.592     \\       
    \hline
    \textbf{Probabilistic combination} &\textbf{0.087} &\textbf{0.659} &\textbf{4.096} \\     
    \bottomrule[1pt]
  \end{tabular}}
  \caption{
    Performance comparison with respect to different combination strategies.
  }
  \label{ablation:combination}
\end{table}

\section{Conclusion}
In this paper, we have revisited the problem of self-supervised monocular depth estimation.
We introduced a simple yet effective method, SQLdepth, in which we 
build a self-cost volume by coarse-grained queries, 
extract depth bins using plane-wise depth counting, 
and estimate depth map using probabilistic combination.
SQLdepth attains remarkable SOTA results on KITTI, Cityscapes and Make3D datasets.
Furthermore, we demonstrate the improved generalization of our model.

\nocite{*}
{\small
\bibliographystyle{ieee_fullname}
\bibliography{egbib}
}

\clearpage
\clearpage
\begin{appendices}

\section{Evaluation Metrics}
We use the standard $7$ metrics to\cite{godard2019digging} quantify the performance
of depth estimators.
Firstly, the $4$ error metrics are defined as:
1) Average relative error (AbsRel): 
$$
    \frac{1}{n} \sum_p^n \frac{\left|y_p-\hat{y}_p\right|}{y}
    $$
2) Root mean squared error (RMSE): 
$$
    \sqrt{\left.\frac{1}{n} \sum_p^n\left(y_p-\hat{y}_p\right)^2\right)}
    $$
3) Squared Relative Difference (SqRel): 
$$
    \frac{1}{n} \sum_p^n \frac{\left\|y_p-\hat{y}_p\right\|^2}{y}
    $$
4) RMSE log: 
$$
    \sqrt{\frac{1}{n} \sum_p^n\left\|\log y_p-\log \hat{y}_p\right\|^2}
    $$
In addition, 3 threshold accuracy metrics are as follows:
$\left(\delta_i\right)$ : percentage of of $y_p$ s.t. 
  $$\max \left(\frac{y_p}{\hat{y}_p},
\frac{\hat{y}_p}{y_p}\right)=\delta<thr 
  $$
for $thr= 1.25, 1.25^2, 1.25^3$, 
where $y_p$ is a pixel in depth map $y$, $\hat{y}_p$ is a pixel in the
predicted depth map $\hat{y}$, and $n$ is the total number of pixels for each
depth map.

\begin{figure}[h]
  \setlength{\columnsep}{0.5pt} 
\begin{multicols}{4}
  \centering

  \begin{subfigure}[b]{0.1175\textwidth}
    \begin{minipage}[t]{\textwidth}
      \centering
      \caption{RGB}
      \includegraphics[width=\textwidth]{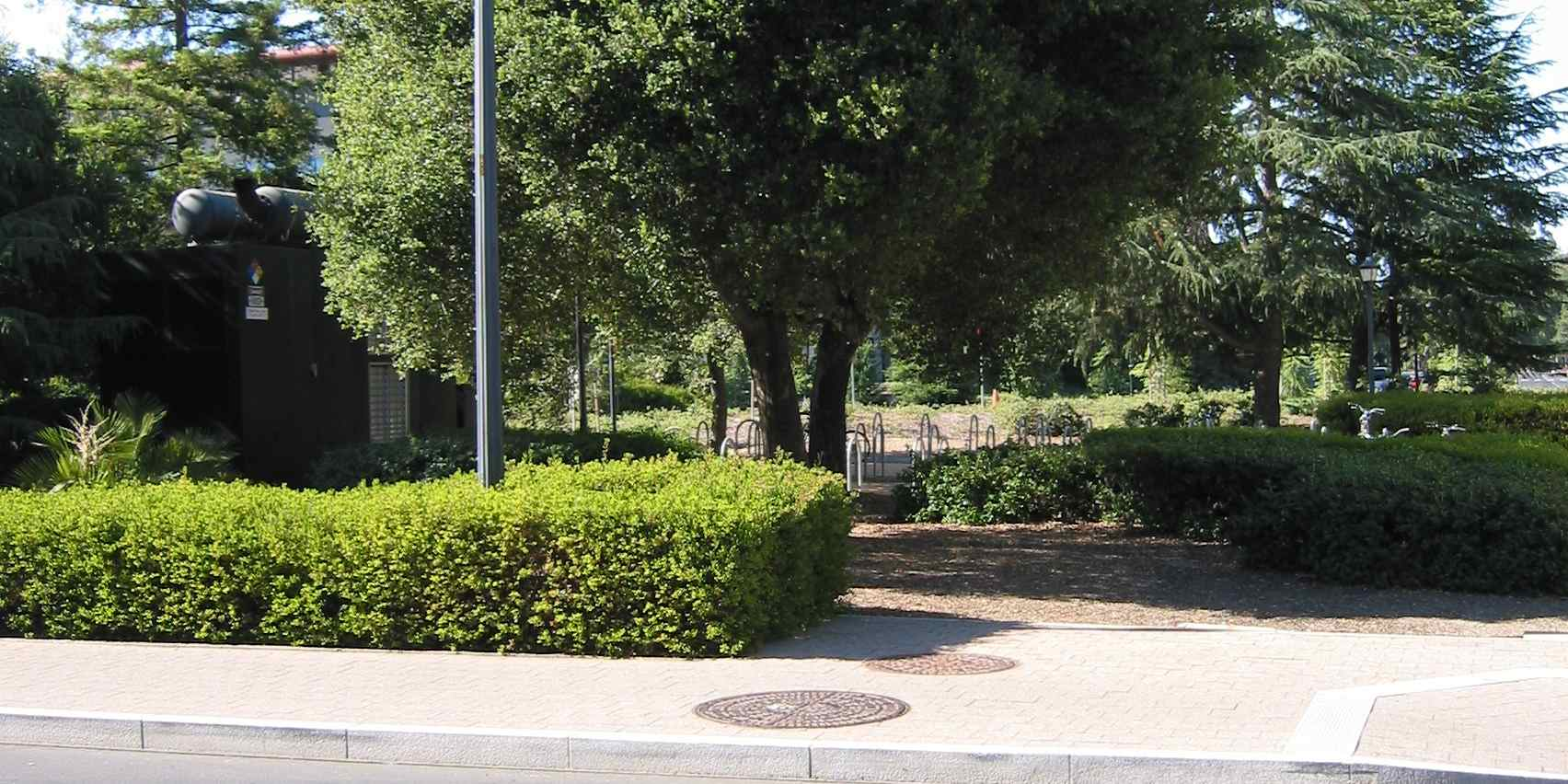}
      \includegraphics[width=\textwidth]{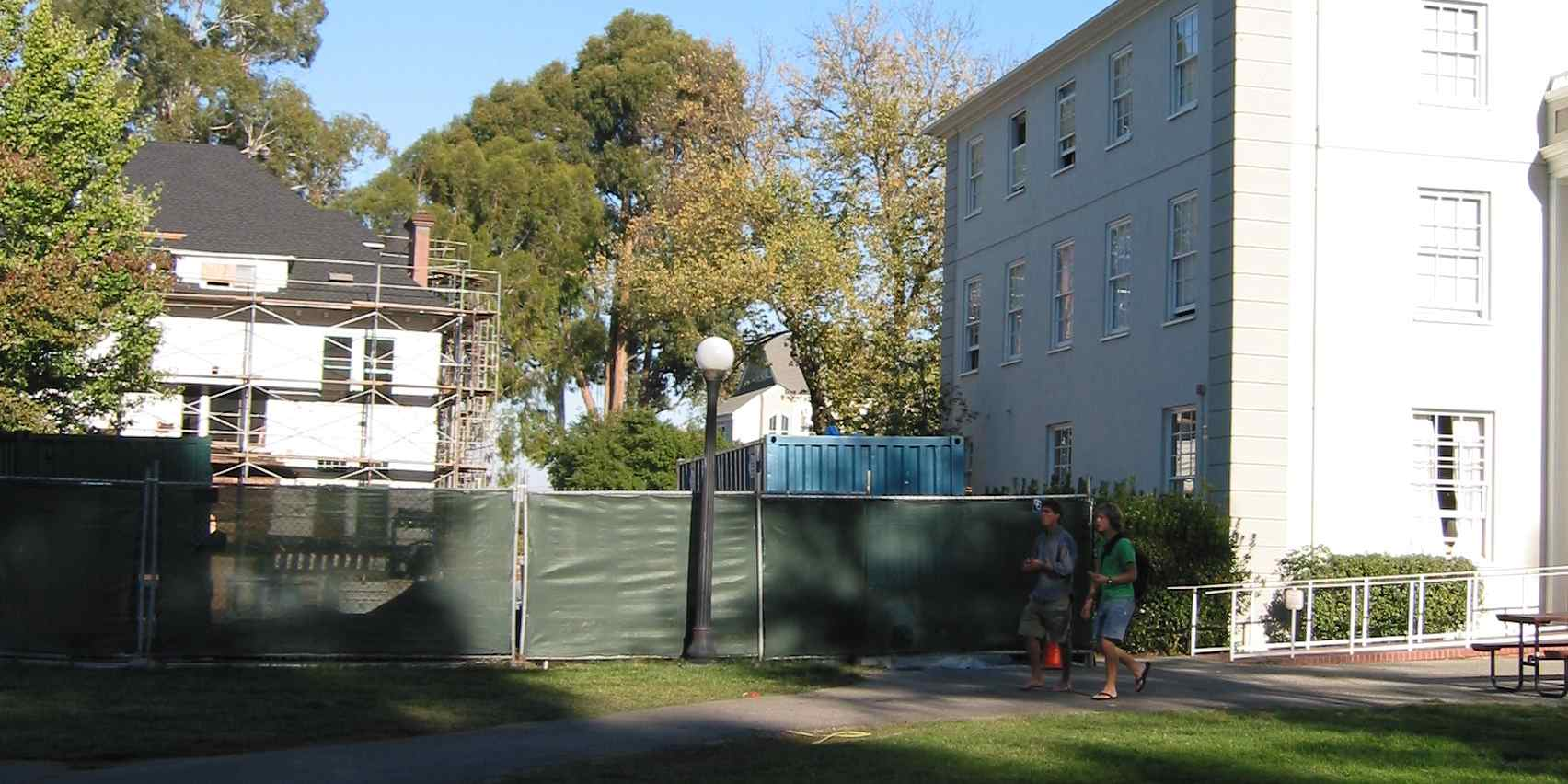}
      \includegraphics[width=\textwidth]{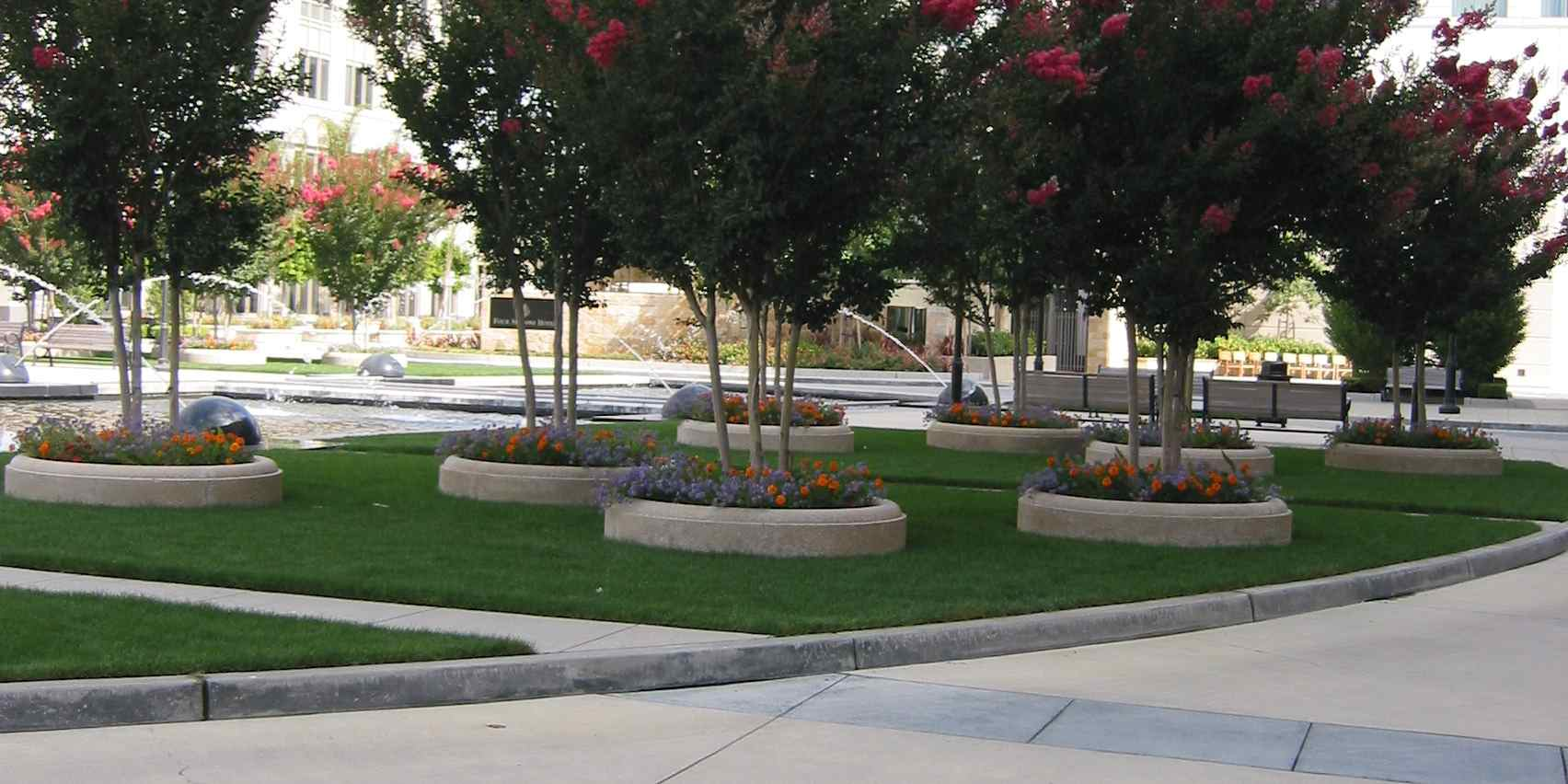}
      \includegraphics[width=\textwidth]{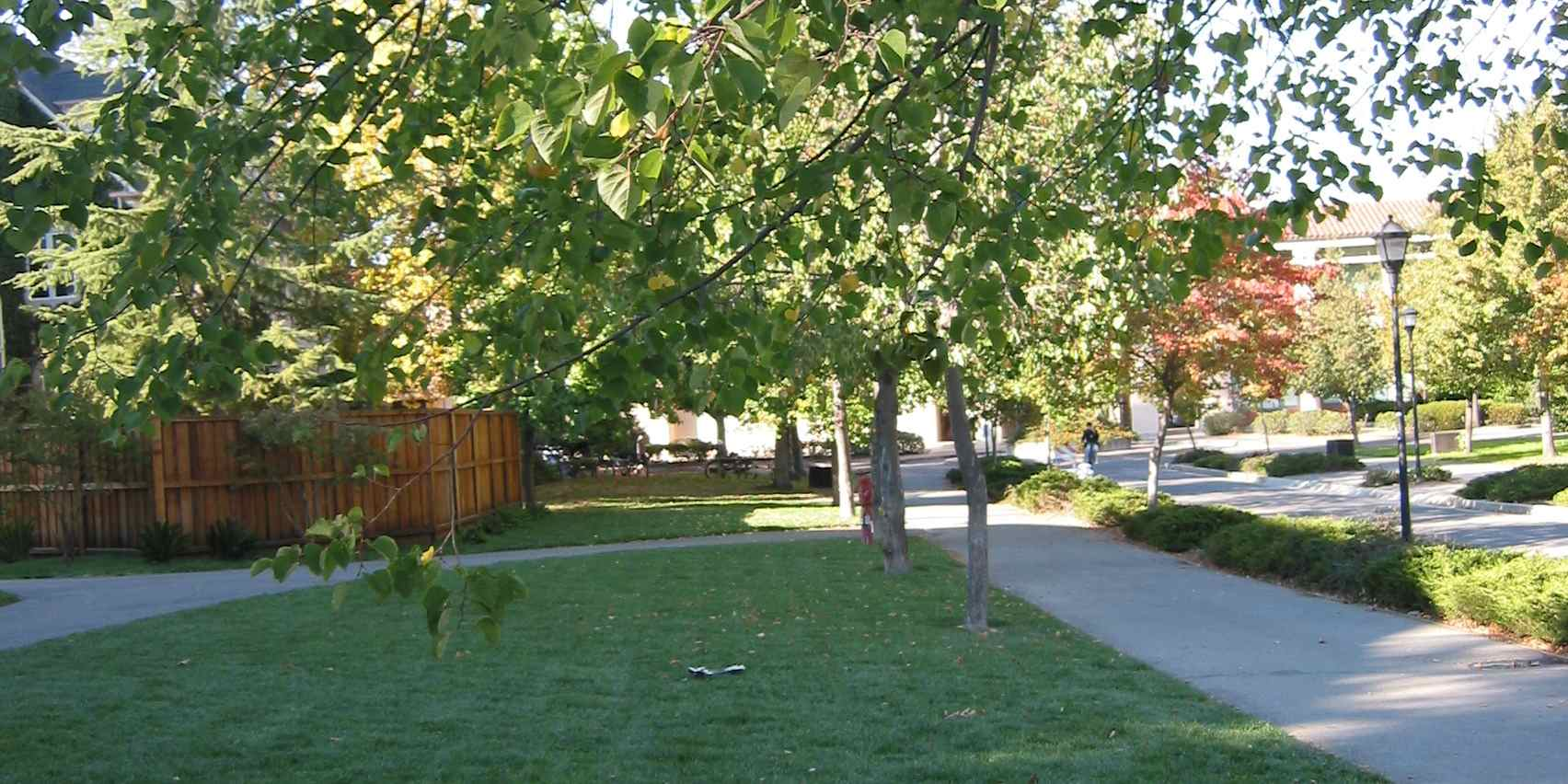}
      \includegraphics[width=\textwidth]{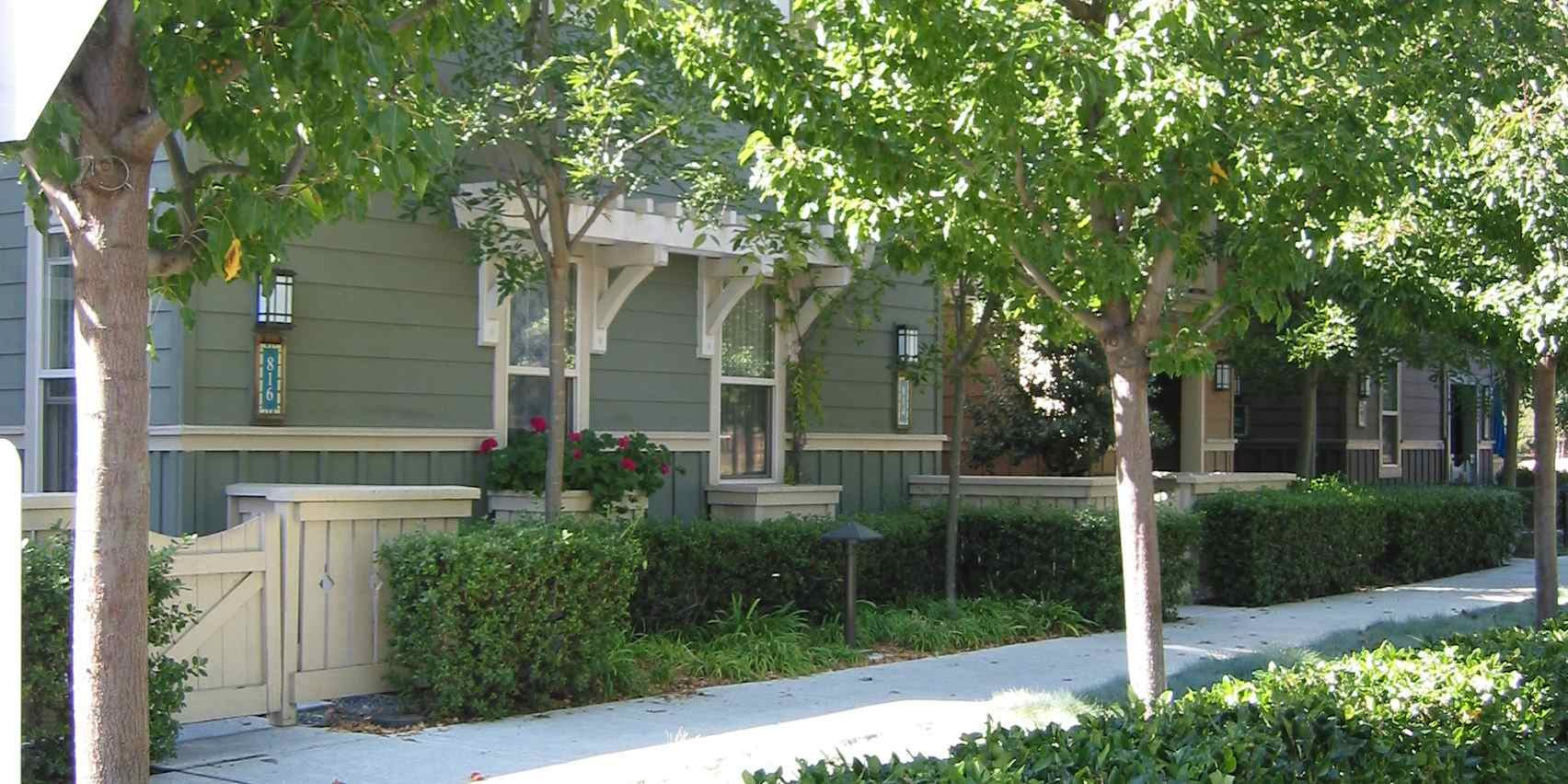}
    \end{minipage}%
  \end{subfigure}

  \begin{subfigure}[b]{0.1175\textwidth}
    \begin{minipage}[t]{\textwidth}
      \centering
      \caption{MD2 \cite{godard2019digging}}
      \includegraphics[width=\textwidth]{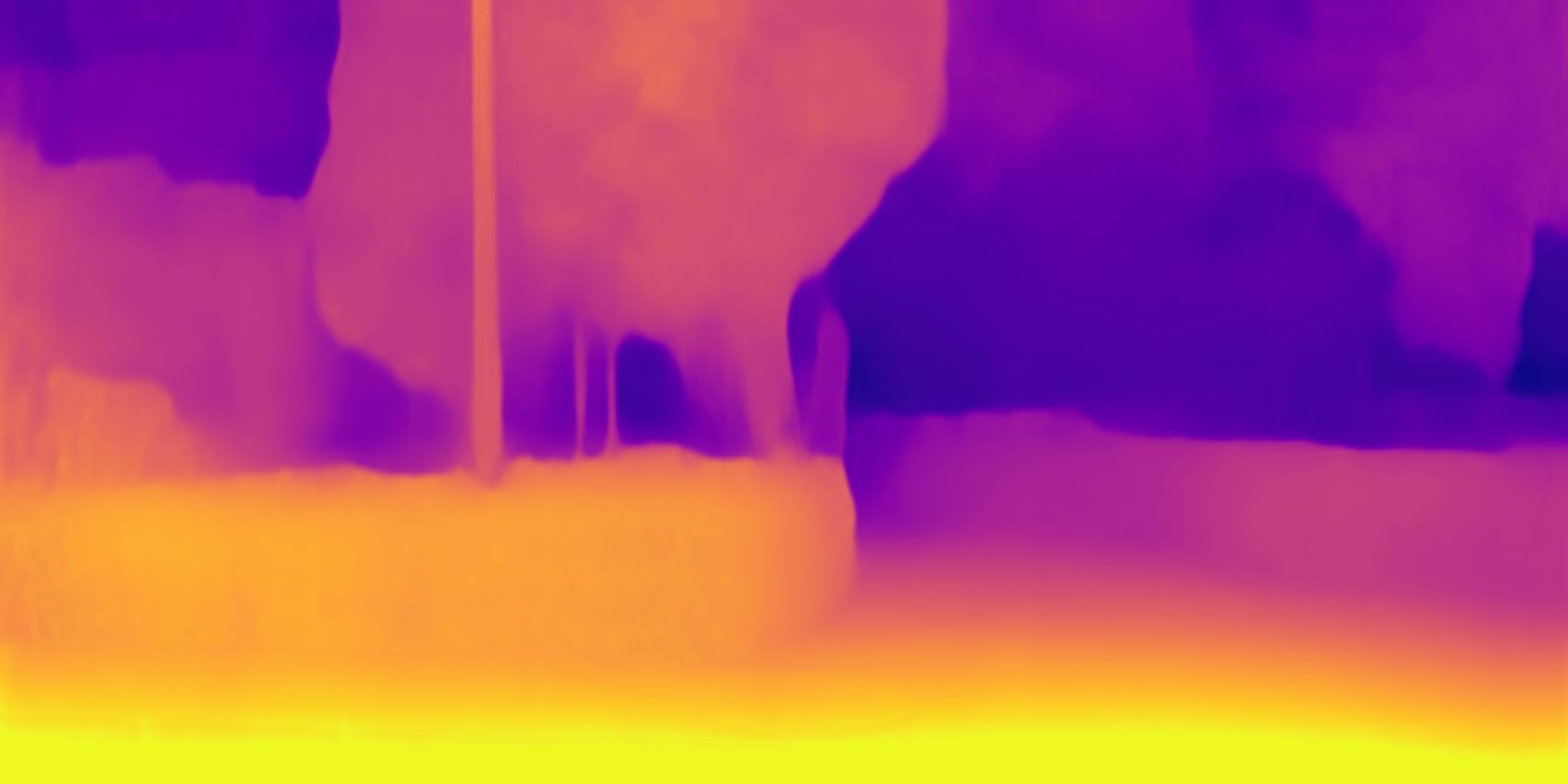}
      \includegraphics[width=\textwidth]{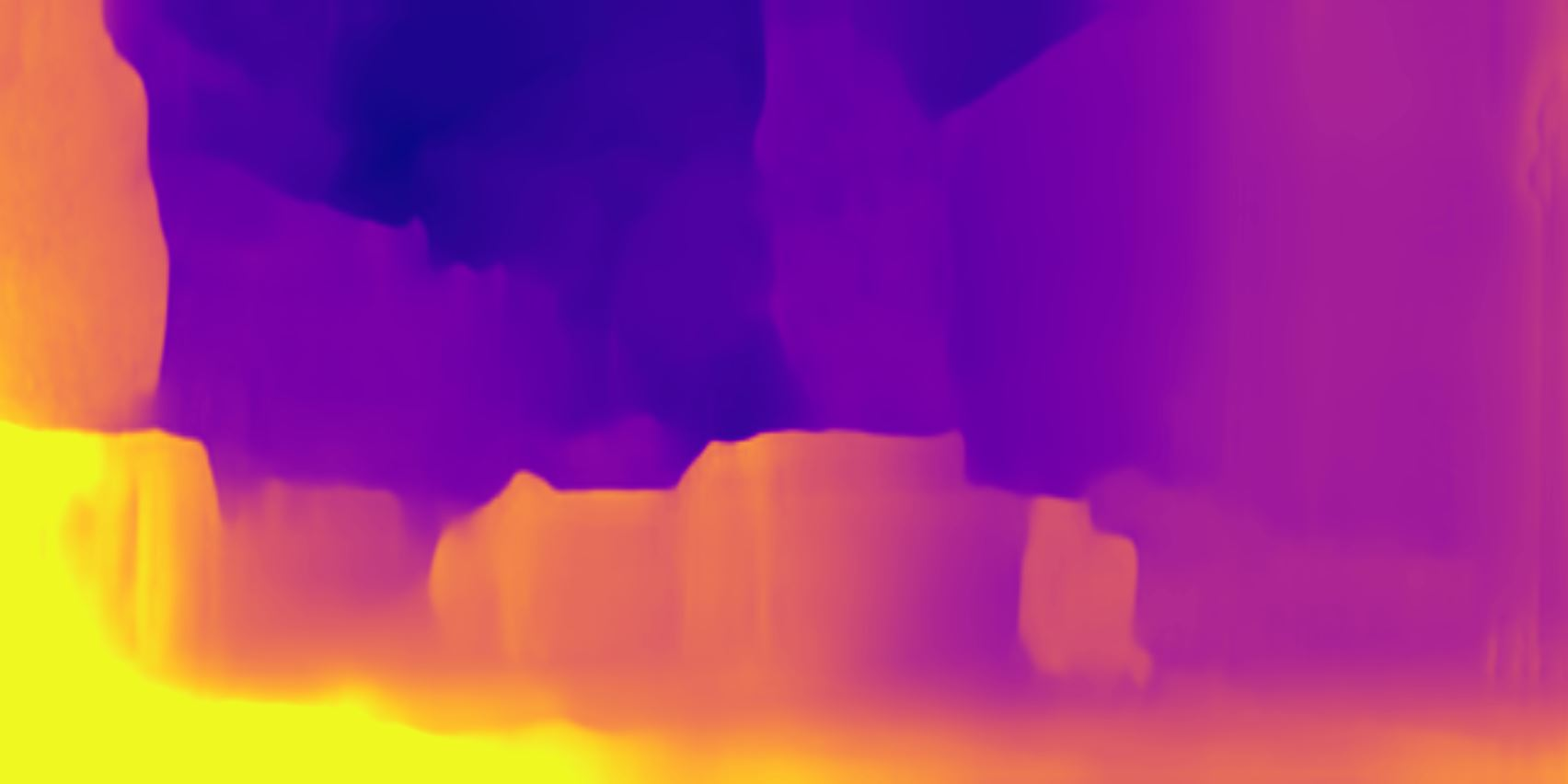}
      \includegraphics[width=\textwidth]{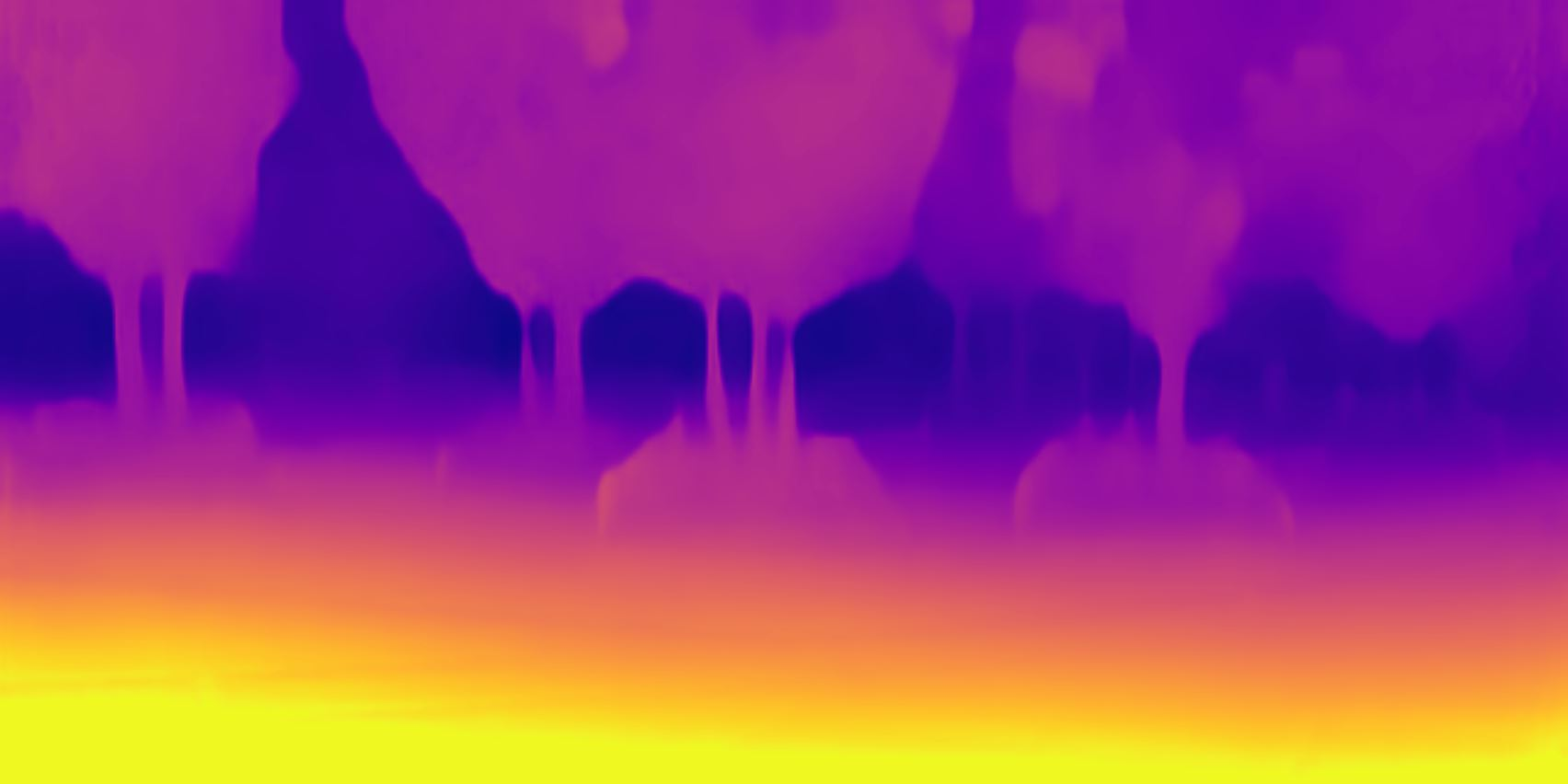}
      \includegraphics[width=\textwidth]{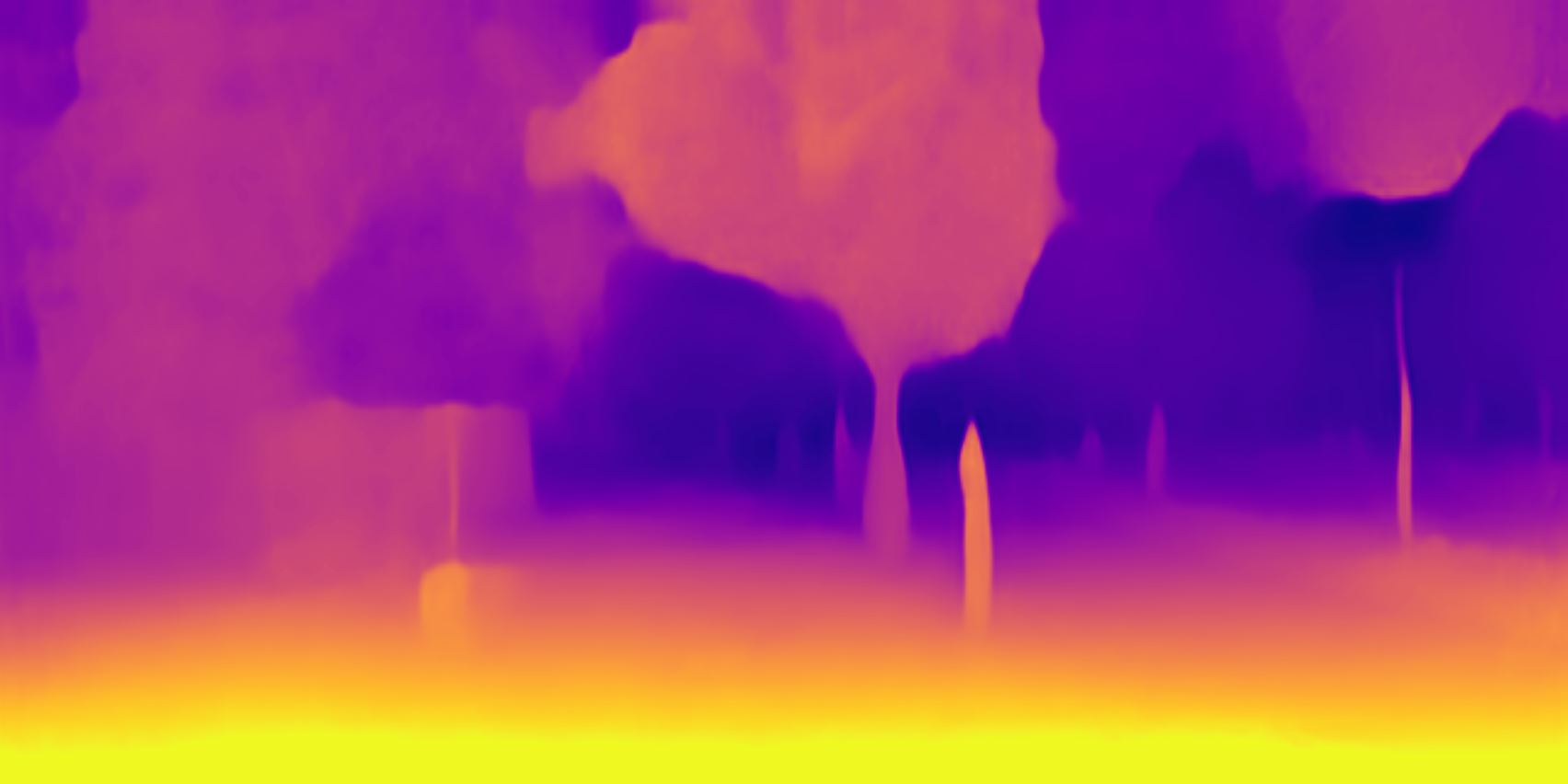}
      \includegraphics[width=\textwidth]{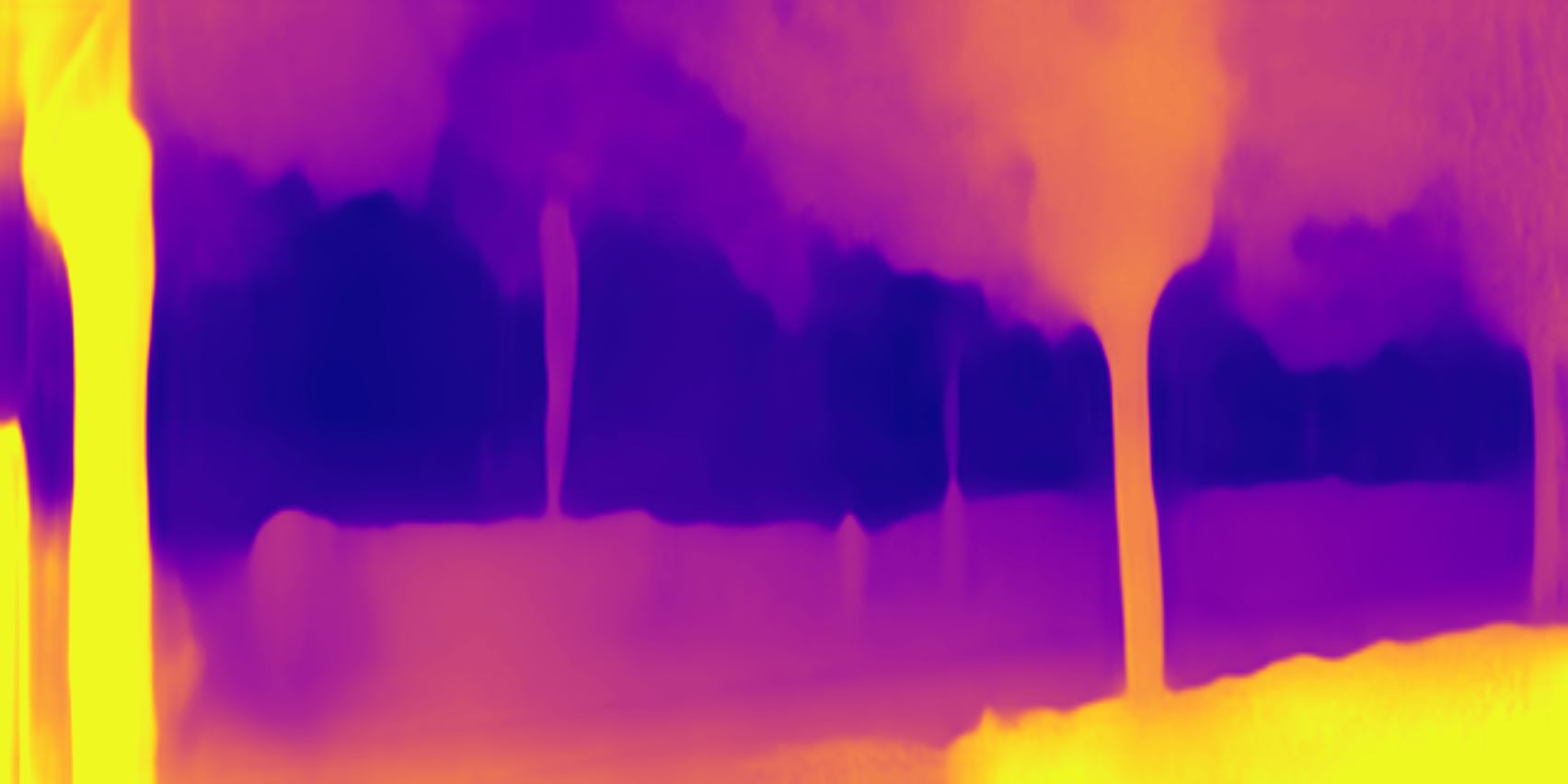}
    \end{minipage}%
  \end{subfigure}

  \begin{subfigure}[b]{0.1175\textwidth}
    \begin{minipage}[t]{\textwidth}
      \centering
      \caption{Ours}
      \includegraphics[width=\textwidth]{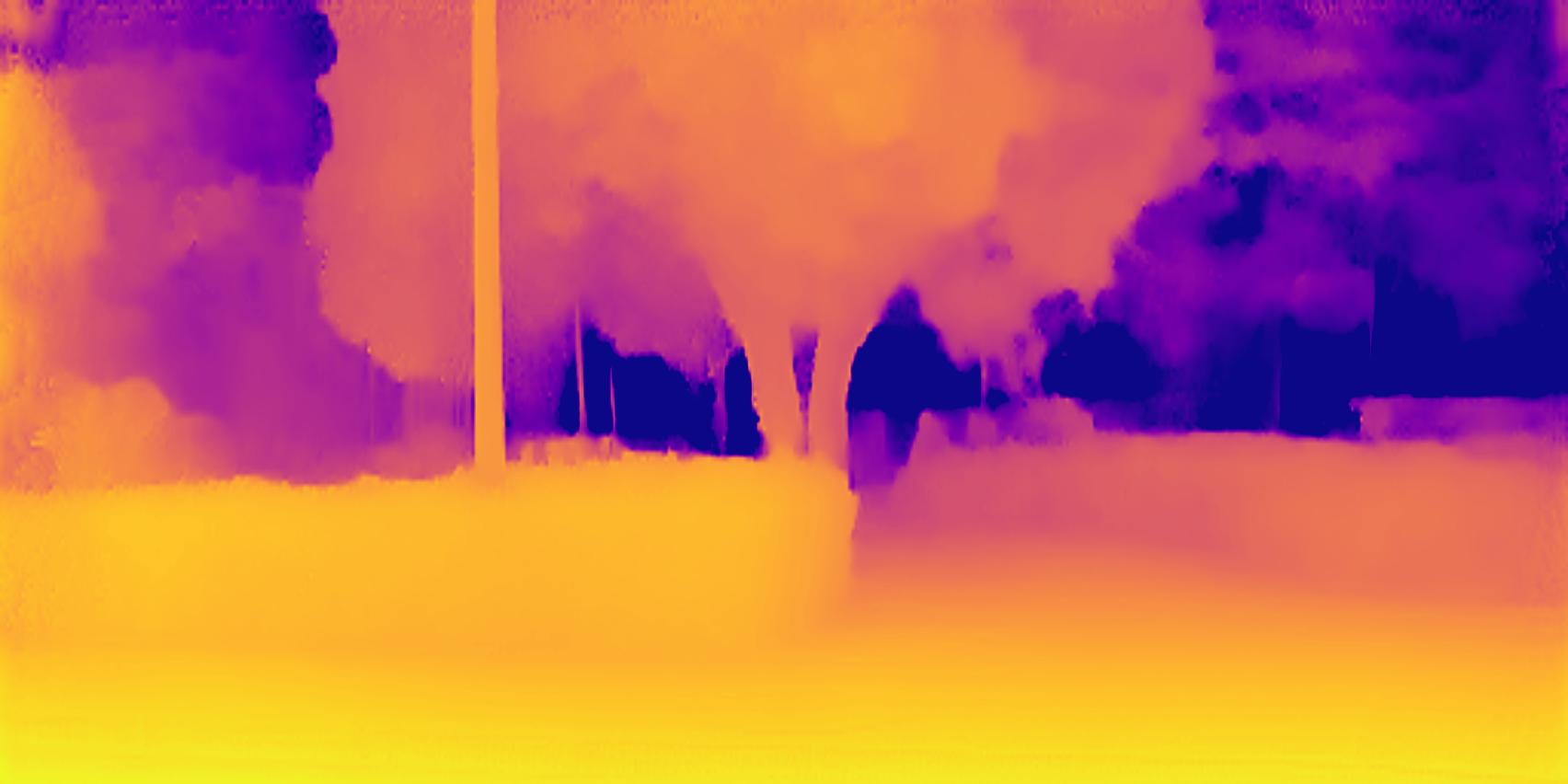}
      \includegraphics[width=\textwidth]{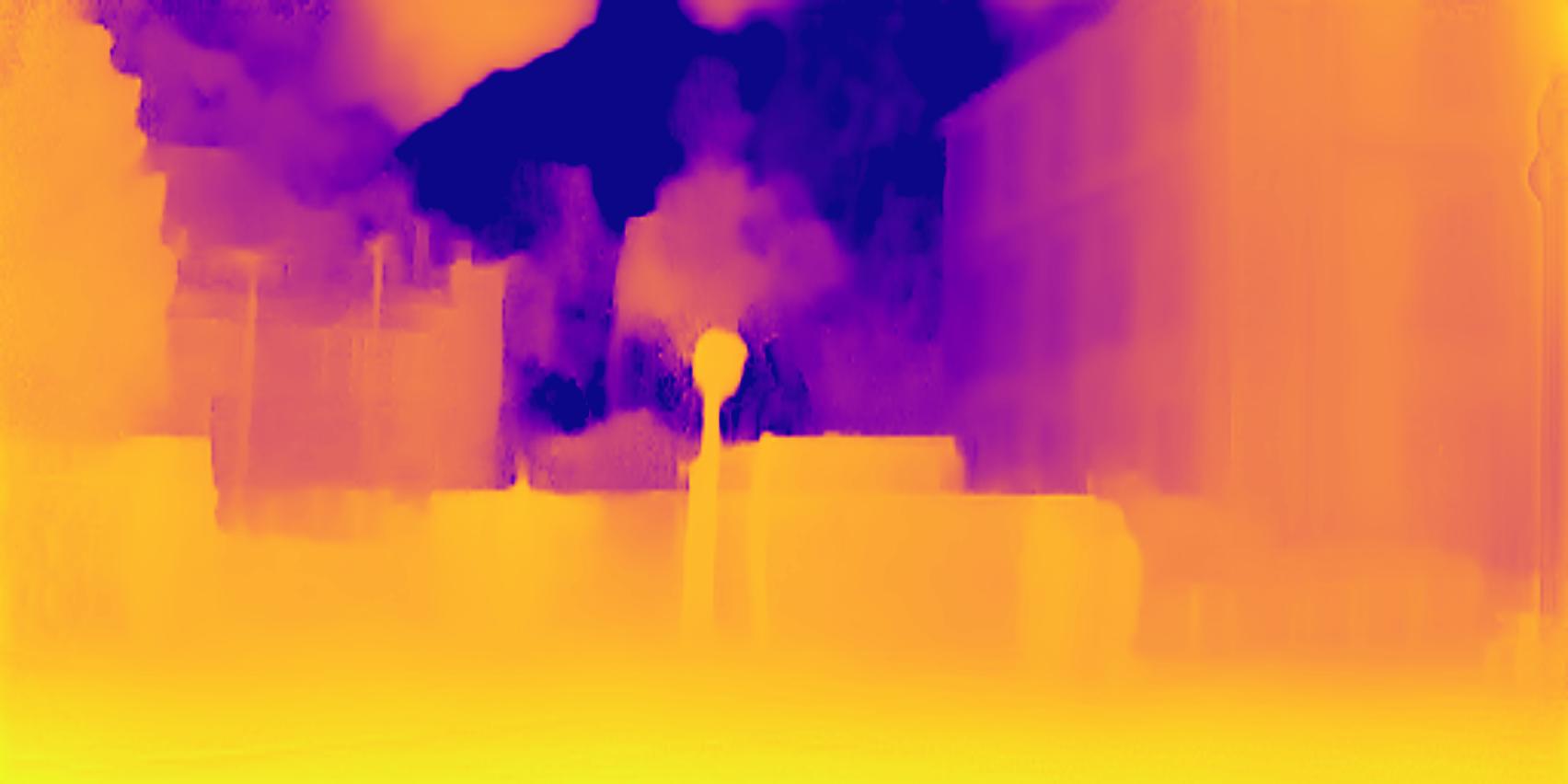}
      \includegraphics[width=\textwidth]{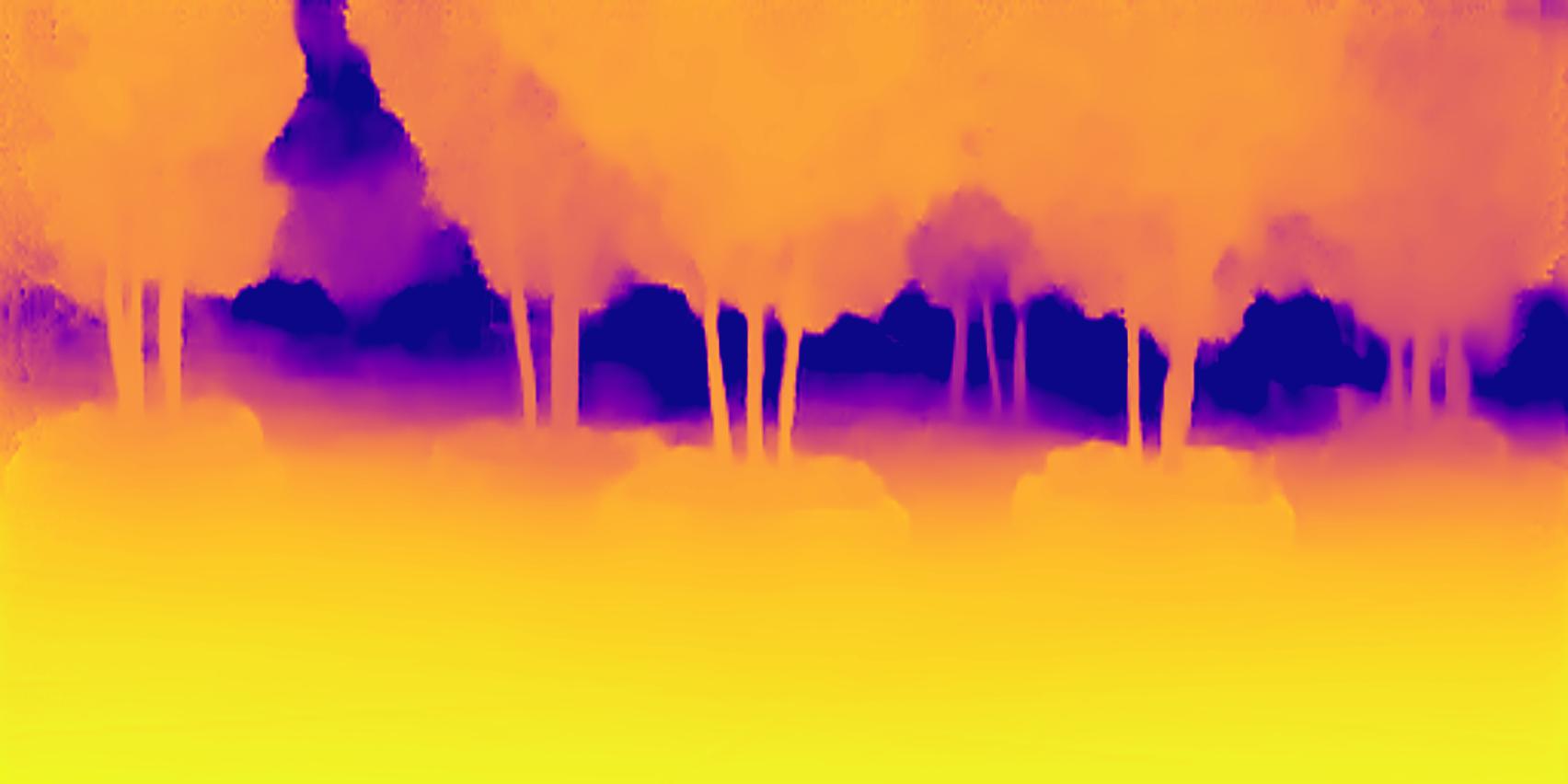}
      \includegraphics[width=\textwidth]{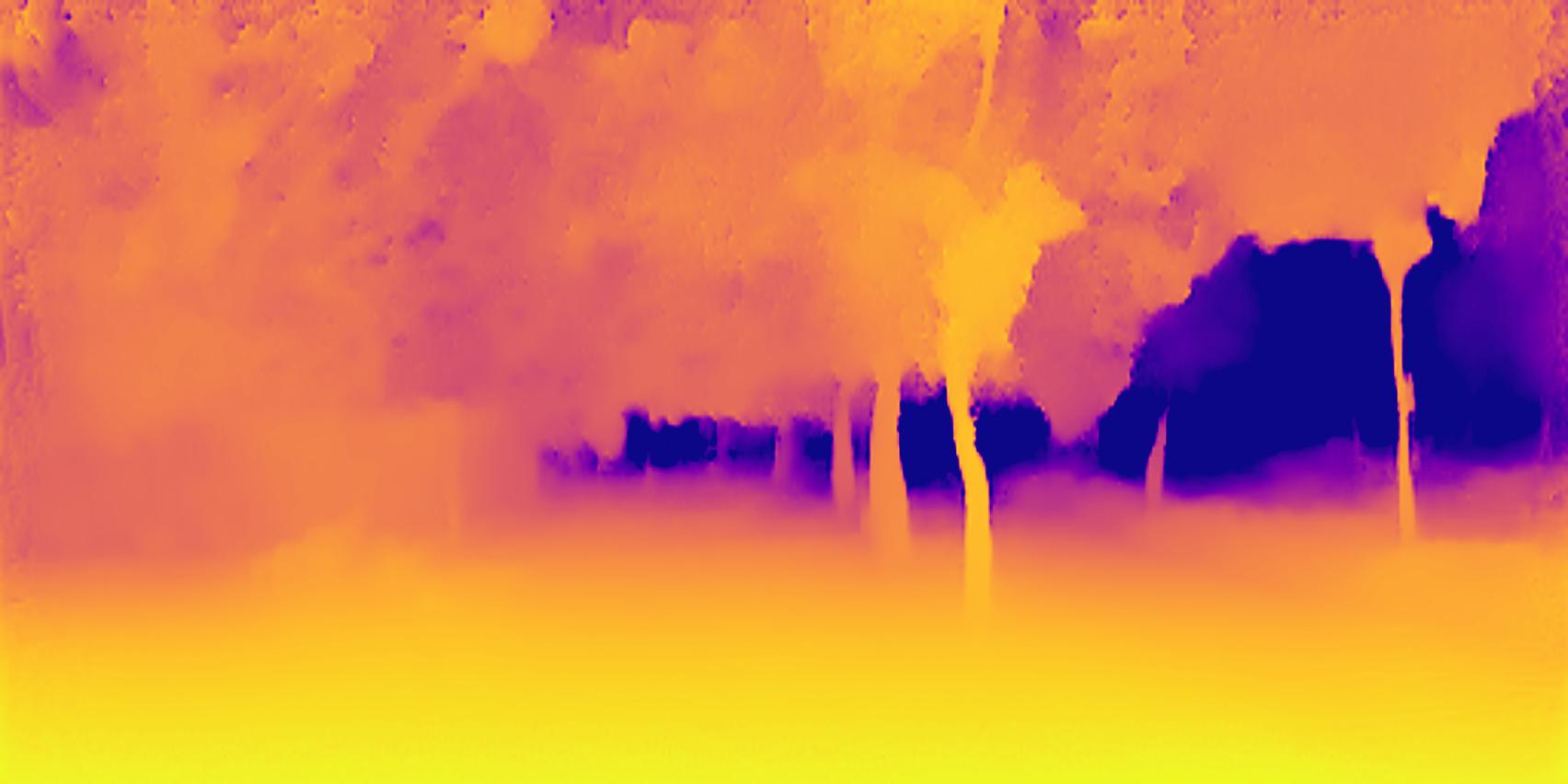}
      \includegraphics[width=\textwidth]{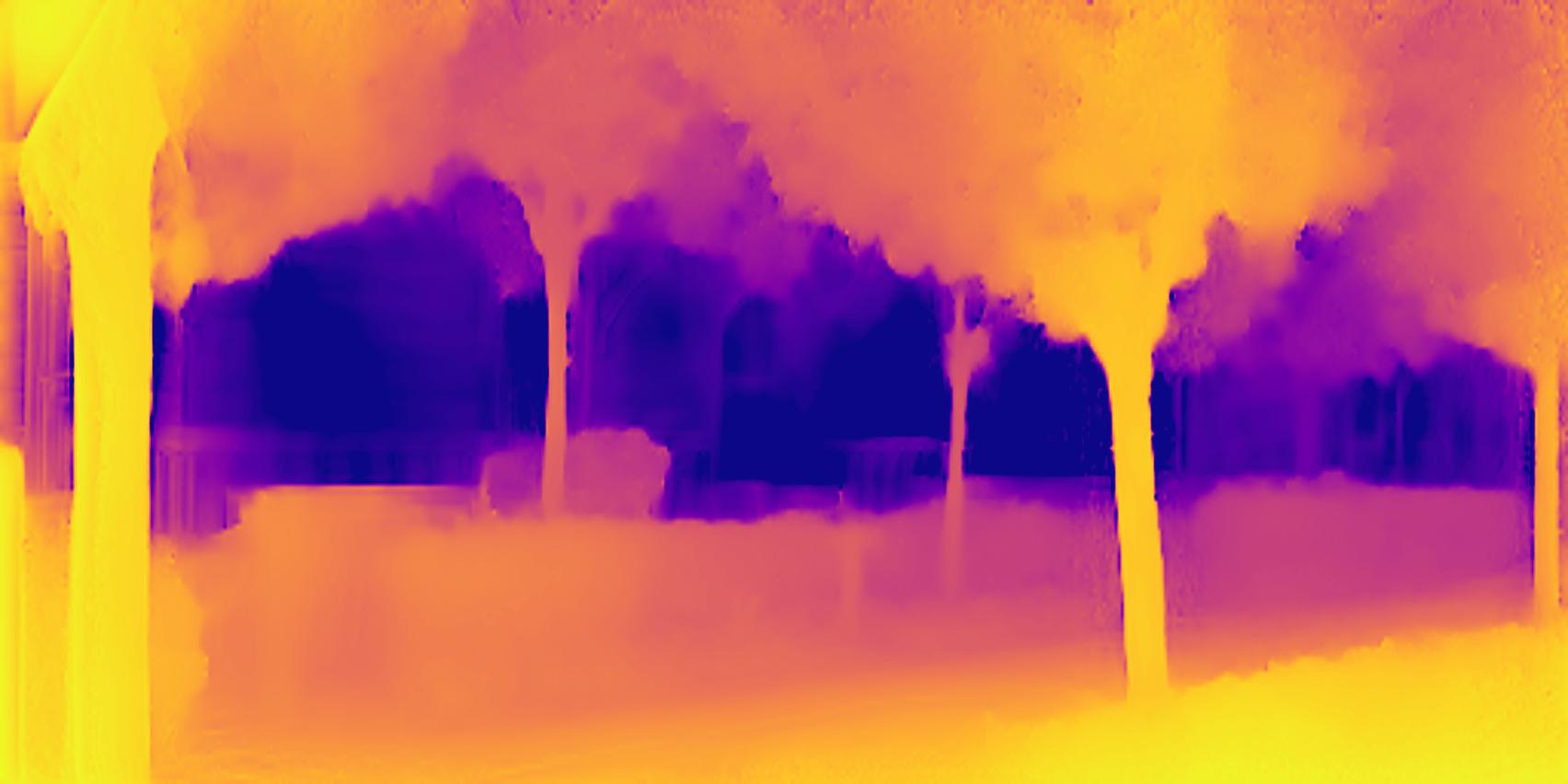}
    \end{minipage}%
  \end{subfigure}

  \begin{subfigure}[b]{0.1175\textwidth}
    \begin{minipage}[t]{\textwidth}
      \centering
      \caption{gt}
      \includegraphics[width=\textwidth]{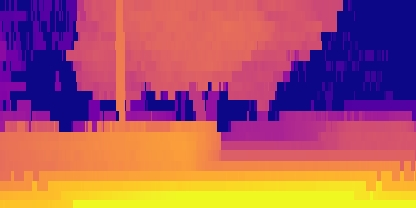}
      \includegraphics[width=\textwidth]{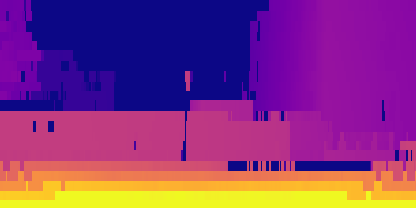}
      \includegraphics[width=\textwidth]{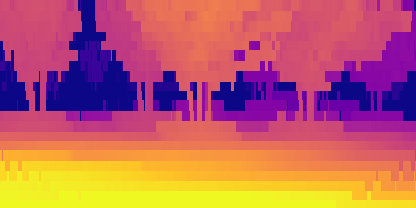}
      \includegraphics[width=\textwidth]{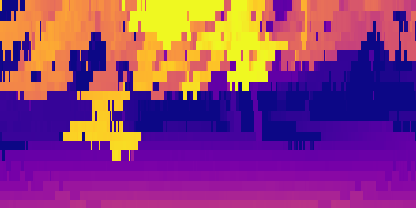}
      \includegraphics[width=\textwidth]{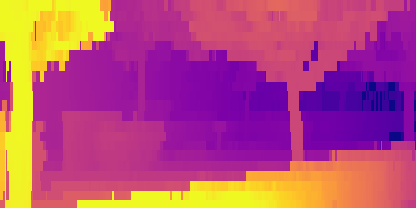}
    \end{minipage}%
  \end{subfigure}

\end{multicols}
  \centering
  \caption{\textbf{More Qualitative Make3D results (Zero-shot)}. 
  }
  \label{fig:Make3D_eval_app}
\end{figure}

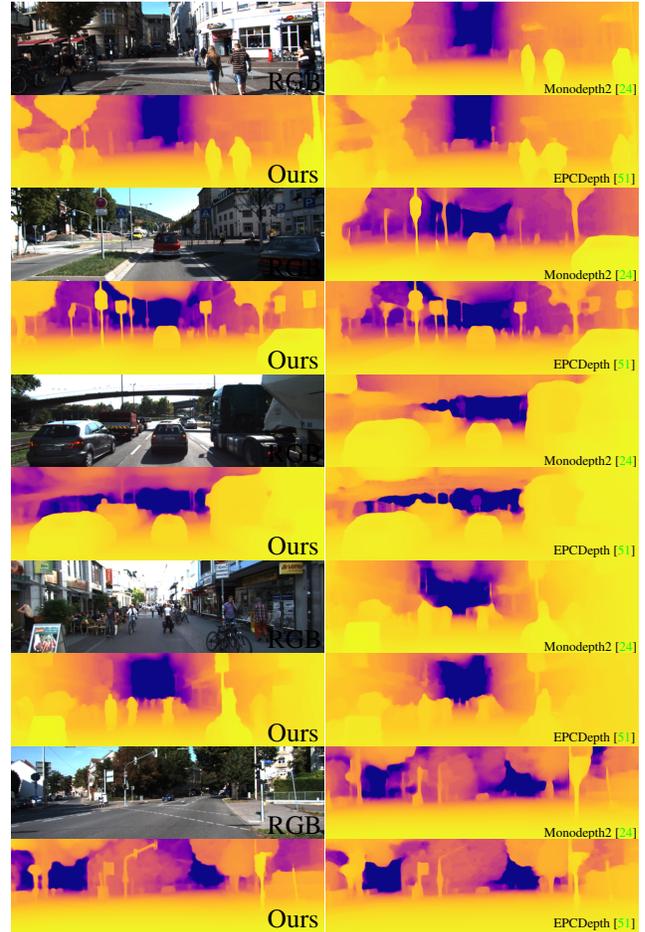
\begin{figure}[t]
  \setlength{\columnsep}{0.5pt} 
\begin{multicols}{2}
  \centering

  \begin{subfigure}[b]{0.25\textwidth}
    \begin{minipage}[t]{\textwidth}
      \centering
      \begin{overpic}[width=0.95\linewidth,keepaspectratio]{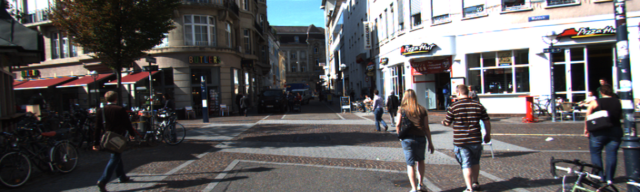}
        \put(82,2){RGB}
      \end{overpic}
      \begin{overpic}[width=0.95\linewidth,keepaspectratio]{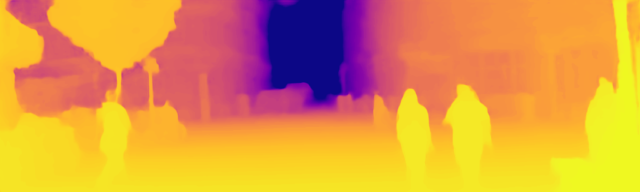}
        \put(82,2){Ours}
      \end{overpic}
      \begin{overpic}[width=0.95\linewidth,keepaspectratio]{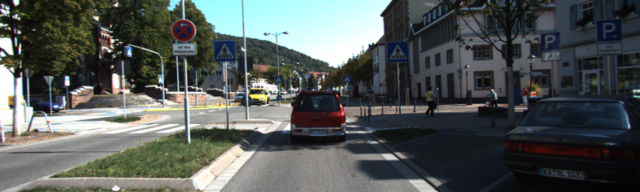}
        \put(82,2){RGB}
      \end{overpic}
      \begin{overpic}[width=0.95\linewidth,keepaspectratio]{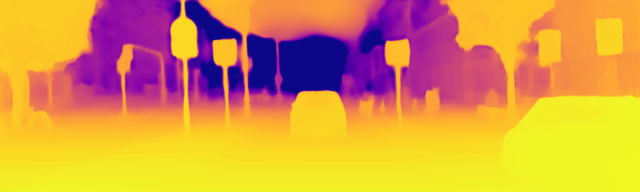}
        \put(82,2){Ours}
      \end{overpic}

      \begin{overpic}[width=0.95\linewidth,keepaspectratio]{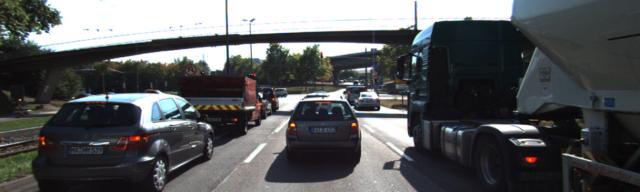}
        \put(82,2){RGB}
      \end{overpic}
      \begin{overpic}[width=0.95\linewidth,keepaspectratio]{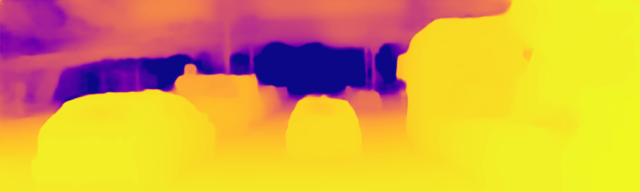}
        \put(82,2){Ours}
      \end{overpic}
      \begin{overpic}[width=0.95\linewidth,keepaspectratio]{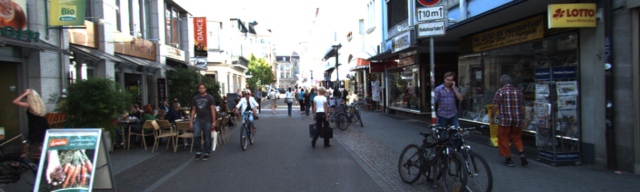}
        \put(82,2){RGB}
      \end{overpic}
      \begin{overpic}[width=0.95\linewidth,keepaspectratio]{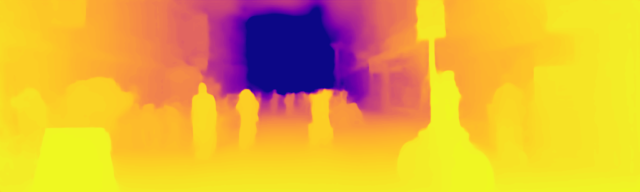}
        \put(82,2){Ours}
      \end{overpic}
      \begin{overpic}[width=0.95\linewidth,keepaspectratio]{./Assets/src_eigen_split_192x640/357.png}
        \put(82,2){RGB}
      \end{overpic}
      \begin{overpic}[width=0.95\linewidth,keepaspectratio]{./Assets/kitti_eval/SQLdepth_unsuper_320x1024_color/357.png}
        \put(82,2){Ours}
      \end{overpic}

    \end{minipage}%
  \end{subfigure}

  \begin{subfigure}[b]{0.25\textwidth}
    \tiny
    \begin{minipage}[t]{\textwidth}
      \centering
      \begin{overpic}[width=0.95\linewidth,keepaspectratio]{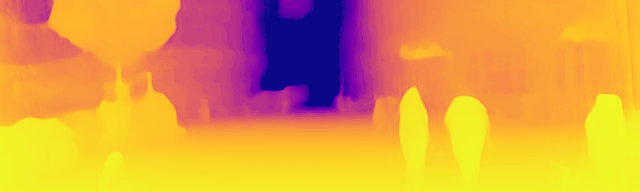}
        \put(70,1){Monodepth2 \cite{godard2019digging}}
      \end{overpic}
      \begin{overpic}[width=0.95\linewidth,keepaspectratio]{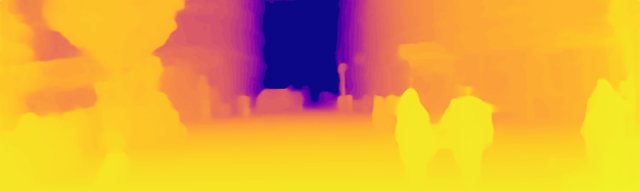}
        \put(73,2){EPCDepth \cite{peng2021excavating}}
      \end{overpic}

      \begin{overpic}[width=0.95\linewidth,keepaspectratio]{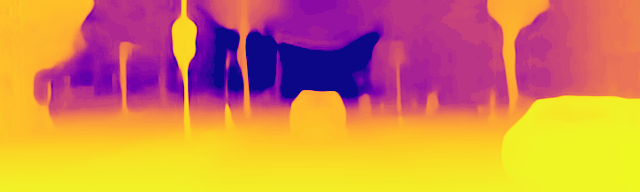}
        \put(70,1){Monodepth2 \cite{godard2019digging}}
      \end{overpic}
      \begin{overpic}[width=0.95\linewidth,keepaspectratio]{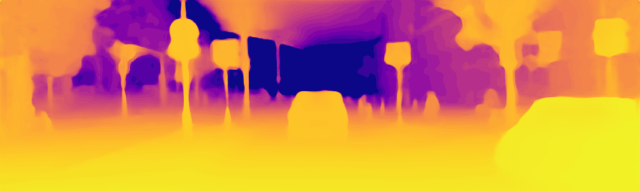}
        \put(73,2){EPCDepth \cite{peng2021excavating}}
      \end{overpic}

      \begin{overpic}[width=0.95\linewidth,keepaspectratio]{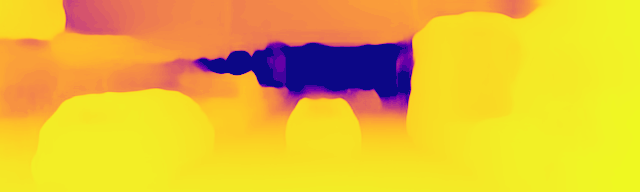}
        \put(70,1){Monodepth2 \cite{godard2019digging}}
      \end{overpic}
      \begin{overpic}[width=0.95\linewidth,keepaspectratio]{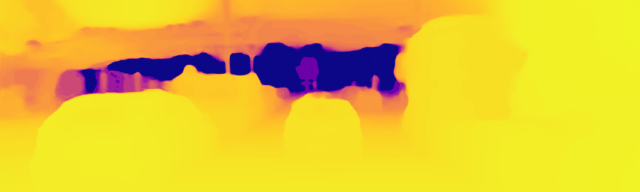}
        \put(73,2){EPCDepth \cite{peng2021excavating}}
      \end{overpic}
      \begin{overpic}[width=0.95\linewidth,keepaspectratio]{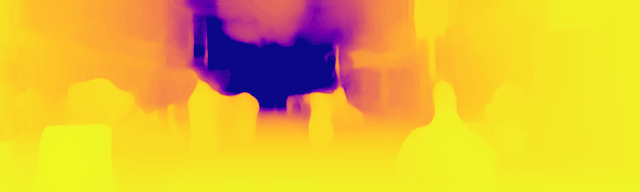}
        \put(70,1){Monodepth2 \cite{godard2019digging}}
      \end{overpic}
      \begin{overpic}[width=0.95\linewidth,keepaspectratio]{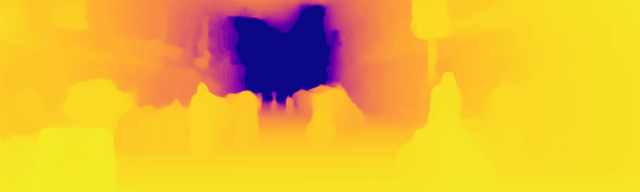}
        \put(73,2){EPCDepth \cite{peng2021excavating}}
      \end{overpic}
      \begin{overpic}[width=0.95\linewidth,keepaspectratio]{./Assets/kitti_eval/mono_640x192_eigen_disp_color/357.png}
        \put(70,1){Monodepth2 \cite{godard2019digging}}
      \end{overpic}
      \begin{overpic}[width=0.95\linewidth,keepaspectratio]{./Assets/kitti_eval/EPC_320x1024_disp_color/357.png}
        \put(73,2){EPCDepth \cite{peng2021excavating}}
      \end{overpic}

    \end{minipage}%
  \end{subfigure}

\end{multicols}
  \caption{\textbf{Additional examples of our predictions on images from KITTI
  dataset.}
  }
  \label{img:addition examples}
\end{figure}

We provide results of ResNet-18 backbone in Table \ref{tab:res18 performance},
and additional comparison with supervised methods in Table \ref{tab: kitti eigen}.

\section{Efficiency and Params Comparison}
In Table \ref{tab:mac performance}, we compared GMACs and params against the full metrics.
The results are calculated with the help of THOP 
\footnote {https://github.com/Lyken17/pytorch-OpCounter \label{foot:flops}}
library.

\begin{table*}[t]
\centering
  \resizebox{\textwidth}{!}{ 
  \begin{tabular}{|c|c|c|c||c|c|c|c|c|c|c|}
\hline
  Method &Train &Test &HxW &\cellcolor{red!25}$AbsRel\downarrow$ &\cellcolor{red!25}$SqRel\downarrow$ &\cellcolor{red!25}$RMSE\downarrow$ &\cellcolor{red!25}$RMESlog\downarrow$ &\cellcolor{blue!25}$\delta<1.25\uparrow$ &\cellcolor{blue!25}$\delta<1.25^2\uparrow$ &\cellcolor{blue!25}$\delta<1.25^3\uparrow$   \\ \hline
\hline
    Ranjan \textit{et al.} \cite{ranjan2019competitive} &M & 1 & 832 x 256 & 0.148 & 1.149 & 5.464 & 0.226 & 0.815 & 0.935 & 0.973 \\
    EPC++  \cite{luo2019every} &M & 1 & 832 x 256 & 0.141 & 1.029 & 5.350 & 0.216 & 0.816 & 0.941 & 0.976 \\
    Struct2depth (M)  \cite{casser2019depth} &M & 1 & 416 x 128 & 0.141 & 1.026 & 5.291 & 0.215 & 0.816 & 0.945 & 0.979 \\
    Videos in the wild  \cite{gordon2019depth} &M & 1 & 416 x 128 & 0.128 & 0.959 & 5.230 & 0.212 & 0.845 & 0.947 & 0.976 \\
    PackNet-SfM \cite{guizilini20203d} &M &1 & 640 x 192 & 0.111 & 0.785 & 4.601 & 0.189 & 0.878 & 0.960 & \underline{0.982} \\
    Patil \textit{et al.} \cite{patil2020don} &M & N & 640 x 192 & 0.111 & 0.821 & 4.650 & 0.187 & 0.883 & \underline{0.961} & \underline{0.982} \\
    Johnston \textit{et al.} \cite{johnston2020self} &M & 1 & 640 x 192 & 0.106 & 0.861 & 4.699 & 0.185 & 0.889 & 0.962 & \underline{0.982} \\
    Monodepth2 (ResNet-18) \cite{godard2019digging}  &M &1 &640 × 192 &0.115 &0.903 &4.863 &0.193 &0.877 &0.959 &0.981 \\
    Wang \textit{et al.} \cite{wang2020self} &M &2(-1, 0) & 640 x 192 & 0.106 & 0.799 & 4.662 & 0.187 & 0.889 & \underline{0.961} & \underline{0.982} \\
    CADepth-Net (ResNet-18) \cite{yan2021channel} &M &1 &640 × 192 &\underline{0.105} &\underline{0.769} &\underline{4.535} &\underline{0.181} &\underline{0.892} &\textbf{0.964} &\textbf{0.983} \\
    \rowcolor{gray!25}\textbf{SQLdepth (ResNet-18)} &M &1 & 640 x 192 &   \textbf{0.098}  &   \textbf{0.749}  &   \textbf{4.486}  &   \textbf{0.176}  &   \textbf{0.900}  &   \textbf{0.964}  &   \textbf{0.983} \\ 
   \hline
  \end{tabular}}
  \caption{
    \textbf{Additional results of ResNet-18 backbone on KITTI eigen benchmark.}
    Results of baselines (self-supervised) are from their paper.
  }
  \label{tab:res18 performance}
\end{table*}

\begin{table*}[t]
\centering
  \resizebox{\textwidth}{!}{ 
  \begin{tabular}{|c|c|c|c|c|c||c|c|c|c|c|c|c|}
\hline
    Method &MACs &Params &Train &Test &HxW &\cellcolor{red!25}$AbsRel\downarrow$ &\cellcolor{red!25}$SqRel\downarrow$ &\cellcolor{red!25}$RMSE\downarrow$ &\cellcolor{red!25}$RMESlog\downarrow$ &\cellcolor{blue!25}$\delta<1.25\uparrow$ &\cellcolor{blue!25}$\delta<1.25^2\uparrow$ &\cellcolor{blue!25}$\delta<1.25^3\uparrow$   \\ \hline
\hline
    Struct2depth (M)  \cite{casser2019depth} &3.5 &14M &M &1 & 416 x 128 & 0.141 & 1.026 & 5.291 & 0.215 & 0.816 & 0.945 & 0.979 \\
    ManyDepth (ResNet-18) \cite{watson2021temporal} &8.3 &20.79M &M &1 & 640 x 192 &0.118 &0.892 &4.764 &0.192 &0.871 &0.959 &0.982 \\
    Monodepth2 (ResNet-18) \cite{godard2019digging}  &8.0 &14.84M &M &1 &640 × 192 &0.115 &0.903 &4.863 &0.193 &0.877 &0.959 &0.981 \\
    Patil \textit{et al.} \cite{patil2020don} &8.0 &14.84M &M &N & 640 x 192 & 0.111 & 0.821 & 4.650 & 0.187 & 0.883 & 0.961 & 0.982 \\
    Wang \textit{et al.} (ResNet-18) \cite{wang2020self} &8.4 &16.78M &M &2(-1, 0) & 640 x 192 & 0.106 & 0.799 & 4.662 & 0.187 & 0.889 & 0.961 & 0.982 \\
    \rowcolor{gray!25}\textbf{SQLdepth  (ResNet-18)} &8.1 &16.2M &M &1 & 640 x 192 &   \textbf{0.098}  &   \textbf{0.749}  &   \textbf{4.486}  &   \textbf{0.176}  &   \textbf{0.900}  &   \textbf{0.964}  &   \textbf{0.983} \\ 
    \hline \hline
    PackNet-SfM \cite{guizilini20203d} &205.49 &128.29M &M &1 & 640 x 192 & 0.111 & 0.785 & 4.601 & 0.189 & 0.878 & 0.960 & 0.982 \\
    FeatDepth\cite{shu2020feature} &75.8 &76.42M &M &1 &1024 × 320 &0.104 &0.729 &4.481 &0.179 &0.893 &0.965 &\underline{0.984} \\
    Monodepth2 (ResNet-50) \cite{godard2019digging}  &29.73 &34.26M &MS &1 &1024 × 320 &0.106 &0.806 &4.630 &0.193 &0.876 &0.958 &0.980 \\
    EPCDepth (ResNet-50) \cite{peng2021excavating}  &48.81 &50.64M &S+Distill &1 &1024 x 320  & 0.091 & \underline{0.646}	& 4.207 & 0.176 & 0.901	& 0.966	& 0.983 \\
    ManyDepth (ResNet-50) \cite{watson2021temporal} &40.2 &37.36M &M &2(-1, 0)+TTR &1024 x 320 & \underline{0.087} & 0.685 & \underline{4.142} & \underline{0.167} & \underline{0.920} & \underline{0.968} & 0.983 \\
    \rowcolor{gray!25}\textbf{SQLdepth (ResNet-50)} &30.57 &37.24M &MS &1 &1024 x 320 &   \textbf{0.082}  &   \textbf{0.607}  &   \textbf{3.914}  &   \textbf{0.160}  &   \textbf{0.928}  &   \textbf{0.972}  &   \textbf{0.985} \\
    \rowcolor{green!25}\textbf{SQLdepth (ConvNeXt-L)} &392.2 &242.15M &MS &1 &1024 x 320 &   0.075  &   0.539  &   3.722  &   0.156  &   0.937  &   0.973  &   0.985 \\
    \hline \hline
    ManyDepth (ResNet-50) \cite{watson2021temporal} &40.2 &37.36M &M &2(-1, 0) + TTR & 1024 x 320 &0.055 &0.305 &2.945 &0.094 &0.963 &0.992 &0.997 \\
    \rowcolor{gray!25}\textbf{SQLdepth (ResNet-50)} &30.57 &37.24M &MS &1 &1024 x 320 &   \textbf{0.052}  &   \textbf{0.223}  &   \textbf{2.550}  &   \textbf{0.084}  &   \textbf{0.971}  &   \textbf{0.995}  &   \textbf{0.998} \\
    \rowcolor{green!25}\textbf{SQLdepth (ConvNeXt-L)} &392.2 &242.15M &MS &1 &1024 x 320 &   0.044  &   0.176  &   2.357  &   0.075  &   0.979  &   0.996  &   0.999 \\
   \hline
  \end{tabular}}
  \caption{
    \textbf{Efficiency and params comparison on KITTI dataset.} 
    We split methods to top half and bottom half, in order to compare light-weight methods and heavy methods respectively.
    TTR is the test-time refinement used by ManyDepth \cite{watson2021temporal}.
    The last three rows are results on KITTI with improved ground-truth.
  }
  \label{tab:mac performance}
\end{table*}

\begin{table*}[ht]
\centering
  \resizebox{\textwidth}{!}{ 
  \begin{tabular}{|c|c|c|c|c||c|c|c|c|c|c|c|}
\hline
    Method &Params &Train &Test &HxW &\cellcolor{red!25}$AbsRel\downarrow$ &\cellcolor{red!25}$SqRel\downarrow$ &\cellcolor{red!25}$RMSE\downarrow$ &\cellcolor{red!25}$RMESlog\downarrow$ &\cellcolor{blue!25}$\delta<1.25\uparrow$ &\cellcolor{blue!25}$\delta<1.25^2\uparrow$ &\cellcolor{blue!25}$\delta<1.25^3\uparrow$   \\ \hline
\hline
    BTS \cite{BTS}     &47M &Sup &- &376x1241 &0.061 &0.261 &2.834 &0.099 &0.954 &0.992 &0.998 \\
    AdaBins \cite{ShariqFarooqBhat2020AdaBinsDE} &78M &Sup &- &376x1241 &0.058 &0.190 &2.360 &0.088 &0.964 &0.995 &\underline{0.999} \\
    \rowcolor{gray!25}SQLdepth (Effcient-b5) &78M &SSL+Sup  &- &376x1241 &0.058 &0.163 &2.129 &0.082 &0.971 &0.995 &\underline{0.999} \\
    ZoeDepth \cite{bhat2023zoedepth}  &345M &Sup &- &376x1241 &0.057 &0.194 &2.290 &0.091 &0.967 &0.995 &0.999 \\
    NeWCRFs \cite{yuan2022new} &270M &Sup &- &376x1241 &0.052 &0.155 &2.129 &0.079 &0.974 &\underline{0.997} &\underline{0.999} \\
    SwinV2-L 1K-MIM \cite{xie2023revealing} &254M &Sup &- &376x1241 &0.050	&0.139 &1.966 &0.075 &0.977 &\textbf{0.998} &\textbf{1.000} \\
    \rowcolor{gray!25}\textbf{SQLdepth (Effcient-b5)} &78M &SSL+Sup &scaling &376x1241 &   0.044  &   0.128  &   2.009  &   0.070  &   0.981  &   \underline{0.997}  &   \underline{0.999} \\
    \rowcolor{gray!25}\textbf{SQLdepth (ConvNeXt-L)} &242M &SSL+Sup &- &376x1241 &\underline{0.043}  &\underline{0.105}  &\underline{1.698}  &\underline{0.064}  &\underline{0.983}  &   \textbf{0.998}  &   \underline{0.999} \\
    \rowcolor{gray!25}\textbf{SQLdepth (ConvNeXt-L)} &242M &SSL+Sup &scaling &376x1241 &   \textbf{0.035}  &   \textbf{0.092}  &   \textbf{1.650}  &   \textbf{0.057}  &   \textbf{0.989}  &   \textbf{0.998}  &   \underline{0.999} \\
    \hline
  \end{tabular}}
  \caption{
    \textbf{Comparison with supervised methods on KITTI \cite{Geiger2013IJRR} eigen benchmark using self-supervised pretrained and metric fine-tuned model.} Scaling for using median-scaling to get absolute depth scale.
  }
  \label{tab: kitti eigen}
\end{table*}
\end{appendices}
\end{document}